\RequirePackage{fix-cm}

\documentclass[twocolumn, natbib]{svjour3}

\smartqed  %

\usepackage[ngerman,english]{babel}

\usepackage{graphicx}
\graphicspath{{img/}}

\usepackage{lineno}
\usepackage[pdfborder={0 0 0}]{hyperref}
\usepackage{layouts}
\usepackage[caption=false]{subfig}
\usepackage{amsmath}
\usepackage{makecell}
\usepackage{color}
\usepackage{microtype} %

\journalname{PFG – Journal of Photogrammetry, Remote Sensing and Geoinformation Science}

\begin{document}

\title{Design, Implementation, and Evaluation of an External Pose-Tracking System for Underwater Cameras}

\author{Birger Winkel \and
        David Nakath  \and
        Felix Woelk   \and
        Kevin Köser
}

\institute{B. Winkel \at
              GEOMAR Helmholtz Centre for Ocean Research Kiel \\ Wischhofstr. 1-3, 24148 Kiel \\
              \email{bwinkel@geomar.de}           %
           \and
           D. Nakath \at
              Christian-Albrechts-Universität zu Kiel, 24118 Kiel, Germany\\
              \& GEOMAR Helmholtz Centre for Ocean Research Kiel \\ Wischhofstr. 1-3, 24148 Kiel \\
              \email{dnakath@geomar.de}
            \and
             F. Woelk \at
              University of Applied Sciences Kiel \\ Grenzstraße 5, 24149 Kiel\\
              \email{felix.woelk@fh-kiel.de}
            \and
               K. Koeser \at
              Christian-Albrechts-Universität zu Kiel, 24118 Kiel, Germany\\
              \& GEOMAR Helmholtz Centre for Ocean Research Kiel \\ Wischhofstr. 1-3, 24148 Kiel \\
              \email{kkoeser@geomar.de}
}

\date{{\color{red} Please cite: \url{https://doi.org/10.1007/s41064-023-00263-x}}}

\maketitle

\begin{abstract}
In order to advance underwater computer vision and robotics from lab environments and clear water scenarios to the deep dark ocean or murky coastal waters, representative benchmarks and realistic datasets with ground truth information are required.
In particular, determining the camera pose is essential for many underwater robotic or photogrammetric applications and known ground truth is mandatory to evaluate the performance of e.g., simultaneous localization and mapping approaches in such extreme environments.
This paper presents the conception, calibration and implementation of an external reference system for determining the underwater camera pose in real-time.
The approach, based on an HTC Vive tracking system in air, calculates the underwater camera pose by fusing the poses of two controllers tracked above the water surface of a tank.
It is shown that the mean deviation of this approach to an optical marker based reference in air is less than 3 mm and 0.3°.
Finally, the usability of the system for underwater applications is demonstrated.

\keywords{tracking \and pose estimation \and Hand-Eye calibration \and UKF \and underwater vision \and ROS \and ground truth \and HTC Vive}
\end{abstract}

\section{Introduction}

\label{sec:introduction}
\begin{figure*}[!t]
    \begin{center}
        \includegraphics[width=\textwidth]{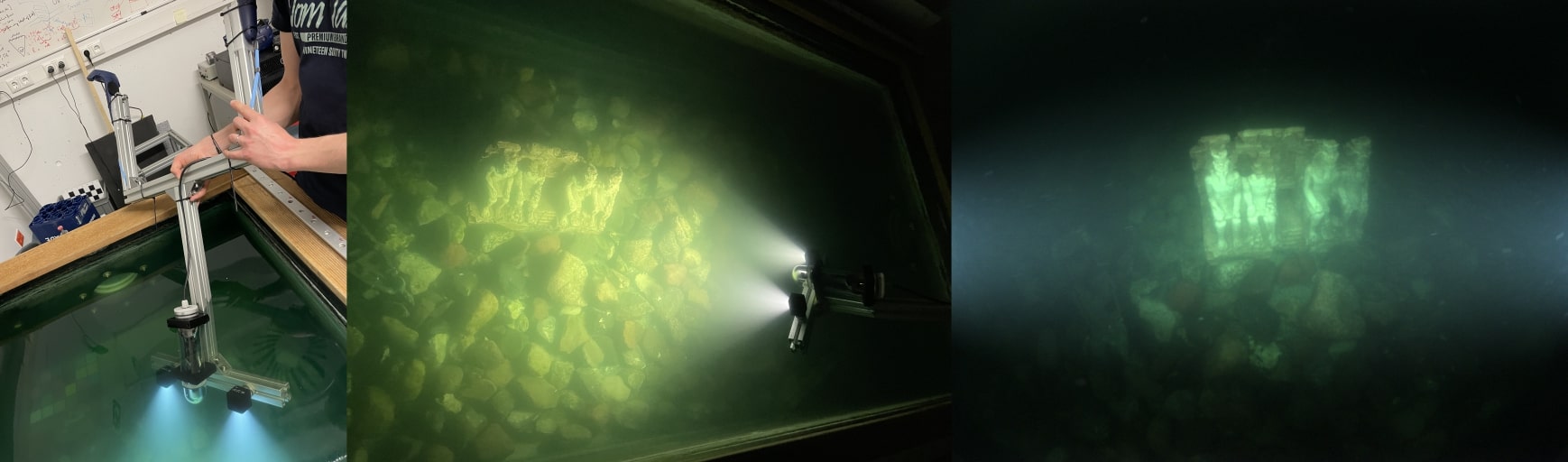}
    \end{center}
    \caption{Application in a deep sea AUV scenario: from left to right: the whole system with the external reference system, underwater camera, and artificial lights; an overview of the scene, without external light sources; and finally a view through the underwater camera.
     \label{fig:teaser} }
\end{figure*}

Pose estimation is a mandatory prerequisite in multiple disciplines, where each case demands the selection of a suitable tracking method.
GNSS (e.g. GPS) for example, is suitable for determining the position of vehicles that have a direct line of sight to multiple satellites, but this method cannot be used underwater due to the attenuation of the signals.
Instead, acoustic methods or optical markers can be used here \citep{surveypaperunderwaternavigation}.
Alternatively, the absolute pose above the water surface can be determined by GNSS and the subsequent relative movements underwater can be measured with an inertial measurement unit (IMU) or the speed relative to the ground with doppler velocity logs (DVLs).
However, these methods can lead to a continuous drift of the pose.
At the same time, the placement of optical markers or other instrumentation at underwater field sites is laborious and can also suffer from challenging environmental conditions such as storms, turbid water, currents and tides, growth of algae or biofouling.
For developing and improving robust visual underwater localization techniques \citep{zhang2022visual} in challenging scenarios \citep{koser2020challenges} such as the deep sea (no sunlight) or turbid waters (coast, harbors), it is difficult to obtain ground truth poses that would allow for the evaluation of underwater computer vision approaches.

In order to robustify and improve such approaches, they should be validated under various test conditions (different water types, different environmental scenarios, different visual structures, different illuminations).
As already argued in \cite{nakath2022optical}, underwater validation scenarios severely suffer from the lack of exactly known conditions, in geometric as well as radiometric terms. 
As an important step to overcome geometric shortcomings, we suggest using a water tank (in our case $2.2 m \times 1 m$ with a depth of $0.8 m$), in which a miniaturized underwater scene is set up that can be seen as a 1:10 model of the real world.
This scene is intended to be used for developing and accessing visual mapping approaches.
To this end, dye and scattering agents are added to the water in the tank to mimic visual impairments known to occur in underwater scenarios (see e.g., \cite{song2022optical} for a comprehensive discussion).
However, the resultant extremely limited visibility makes it difficult to determine ground truth camera poses underwater using optical markers \citep{paperPhilip}.
On top of that, adding optical markers will also create unrealistic structures, potentially biasing an evaluation.
Our approach is therefore to implement an external reference system, which can determine ground truth camera poses independently of the water conditions.
An optical tracking system is used for this, which determines absolute poses above the water surface and applies a rigid transformation and sensor fusion to obtain the camera pose underwater in real time.

An HTC Vive Pro is chosen as the tracking system.
This consumer virtual reality (VR) system \cite{vrBasic} is significantly cheaper than comparable optical motion capture systems such as VICON or OptiTrack.
Another advantage of the HTC Vive is the easy integration of the tracking algorithms into the robot operating system (ROS) \citep{quigley2009ros} using the OpenVR interface.
The ground truth controller poses are thus available in ROS, so that the rigid transformation to the camera pose can be determined for each controller with a Hand-Eye calibration approach.
To reduce the tracking and transformation errors, two controllers are used on the upper end of the camera stick.
This results in two pose estimates for the camera, which are merged using an Unscented Kalman Filter (UKF) \citep{julierNewExtensionKalman1997,wanUnscentedKalmanFilter2000}.
Finally, a custom-made underwater camera-light system, which can additionally be equipped with IMUs, is mounted at the end of the stick to record underwater datasets with ground truth poses, see \autoref{fig:teaser}.

In this work, the tracking quality of the developed system is quantitatively evaluated. 
For this purpose, the tracking and the Hand-Eye calibration are analyzed first.
Then the precision of the real-time camera pose estimation is validated.
This is achieved by first measuring an optical target as a fixed point in space.
Then the camera pose of the external tracking system is compared with a camera pose determined optically under good visibility conditions.

\section{Related work}
\label{sec:related_work}

\paragraph*{HTC Vive analysis}
$\\$
Using the HTC Vive as a cost-effective tracking system in the field of robotics and computer-vision is widespread see e.g., \cite{wang2020digital,ayyalasomayajula2020deep,nasaVive}.
In the latter paper, the accuracy of tracking of first generation HTC Vive devices is analyzed by using the open-source library Libdeepdive \citep{libdeepdive} in combination with a subsequent optimization step.
Trackers in different orientations are used together with two lighthouses.
Being restricted to a 2D plane, a mean standard deviation of 1.18 mm and 0.45° is achieved for the tracker poses across the data sets.
An analysis for the accuracy of the HTC Vive headset tracking is performed in \cite{viveResearch}.
However, since the headset is too big and bulky to use it for our experiments, there are no comparable values for our target application.
Finally, \citep{s21051622} provide and in-depth pose estimation accuracy evaluation of the second generation of the Vive.
However, all the results are not comparable to our approach, as we are interested in fused results of a pose with significant offset.
\paragraph*{Hand-Eye calibration}
$\\$
Hand-Eye calibration describes a technique for calculating the 3D Position and orientation of a camera relative to a robot manipulator.
In this work, it is used to determine the rigid transformation between the controllers and the underwater camera on the camera stick.
Various approaches can be found in the literature (see e.g., \cite{enebuse2021comparative}).
The classic approach is from \cite{tsai}, in which the pose of the hand is known from the robot or an external tracking system.
A calibration target is used here, which is placed firmly in the room.
Then a data set of poses for the hand and the transformation from the camera to the calibration target is recorded.
From this, a linear system of equations can be set up and the rigid transformation between hand and camera can be determined.

An alternative approach does not rely on calibration targets and instead determines the optical data with a structure-from-motion approach.
This technique is particularly interesting for areas of application in which no calibration target can be set up.
In \cite{handEyeStructureFromMotion}, an example application in the medical field is demonstrated, in which a non-sterile calibration pattern cannot be used.
However, the accuracy of this approach is, on average, lower than that of the classical approach, which we hence stick to.

\paragraph*{Underwater ground truth camera poses}
$\\$
Simulated or heavily controlled environments can often help to investigate parts of a complex problem under known conditions.
The idea of hardware in the loop systems is to include real hardware in a simulation system, to narrow the simulation to reality gap \citep{bacic2005hardware}. 

Robot arms are often used as an external reference system for determining camera poses as they offer a high repeatability and thus offer a good ground truth reference \citep{robotCamPose}.
Hence, they are often used in hardware in the loop test facilities for e.g., spacecraft \citep{park2021robotic,kruger2010tron}.

However, a robot arm for the planned application requires a large radius of movement, which increases the acquisition costs significantly.
At the same time, the integration into an underwater application results in a high maintenance effort, so that this approach is not pursued by us.

A GPS-INS fusion-based system is used in \cite{bleier2019scout3d} to estimate the 6DOF of a ship which scans the ground with a laser system. However, as we work in an indoor lab-environment, we do not use satellites.
In \cite{bernal2017development} a VICON tracking system is used in conjunction with immersed trackers and off-the-shelf cameras to capture the motion of a space suit underwater. This approach cannot be employed by us, as we want to operate in extreme visibility conditions, which prohibit easy detection underwater.
Two interesting solutions for cross-surface pose-determination are presented in \cite{nocerino2020photogrammetry}: (i) a rod is equipped with a stereo-camera-system with a submerged as well as an in-air image (ii) a rod can as well be equipped with markers above and below the surface and then be observed by one camera.
Still, we cannot use such approaches, as we also want to model different light conditions, including deep sea environments with artificial lighting, which demand absolutely no light over the surface.

Comparable work using an external reference system for underwater camera applications is presented in \cite{otherUnderwaterPaper}.
Four optical tracking systems (VICON) are used here to detect poses from infrared sensors mounted at the upper end of a rod.
This is a much more expensive system compared to a consumer VR system.
The authors provide no analysis of the accuracy of the motion capture system, but it is claimed that it achieves an accuracy down to 0.5 mm in a $4 m \times 4 m$ volume.
At the bottom of the rod an underwater camera with integrated IMU is located.
For this purpose, the absolute deviation of the trajectory for VINS-Fusion \citep{qinGeneralOptimizationbasedFramework2019} and ORB-SLAM2 \citep{mur-artalORBSLAM2OpenSourceSLAM2017} are compared to the ground truth camera poses over several data sets.
In the case of VINS-Fusion, the deviation is on average 67.63 mm with the IMU and 111.33 mm without its inclusion.
ORB-SLAM2 achieves an average deviation of 88.1 mm.

\section{System design}
\label{sec:system_design}

\subsection{Hardware overview}
\begin{figure}[!t]
    \begin{center}
        \includegraphics[width=.5\textwidth]{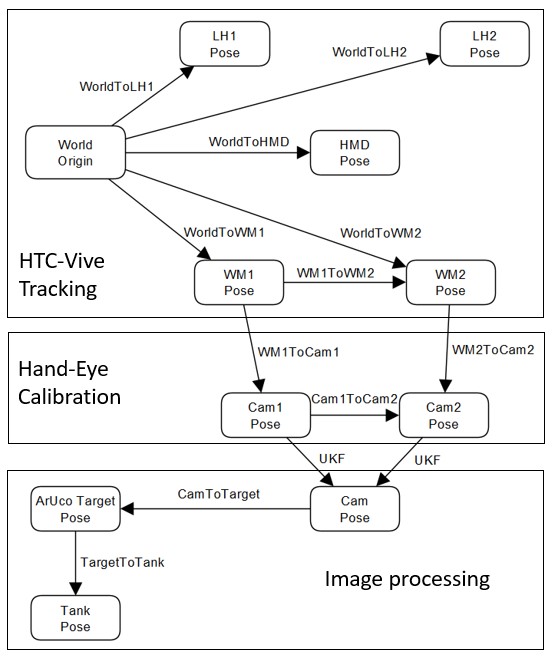}		
    \end{center}
    \caption{Overview of system transformations. \label{fig:surviveWorld} }
\end{figure}

In \autoref{fig:surviveWorld} the connection between the hardware components is shown as a block diagram.
This includes the devices from the HTC Vive, the camera, the calibration target consisting of ArUco markers and the water tank.
The respective poses and the transformations used in 3D space are shown.
The poses of the controllers are determined by the tracking system. 
With the Hand-Eye calibration result, a rigid transformation for each controller to the camera pose is concatenated.
The optimal camera pose is calculated by the UKF, by fusing the two noisy measurements of the controller poses.
Finally, the transformation from the camera to the calibration target can be determined using ArUco marker detection.
By placing the calibration target in a corner of the water tank, its pose can be defined in world coordinates by concatenating the inverse transformations.
For more information about different coordinate systems, we refer to \cite{cooSys}.

\subsection{Vive Setup}
\begin{figure}[!t]
    \begin{center}
        \subfloat[Side view of real environment]{
            \includegraphics[height=0.35\textwidth]{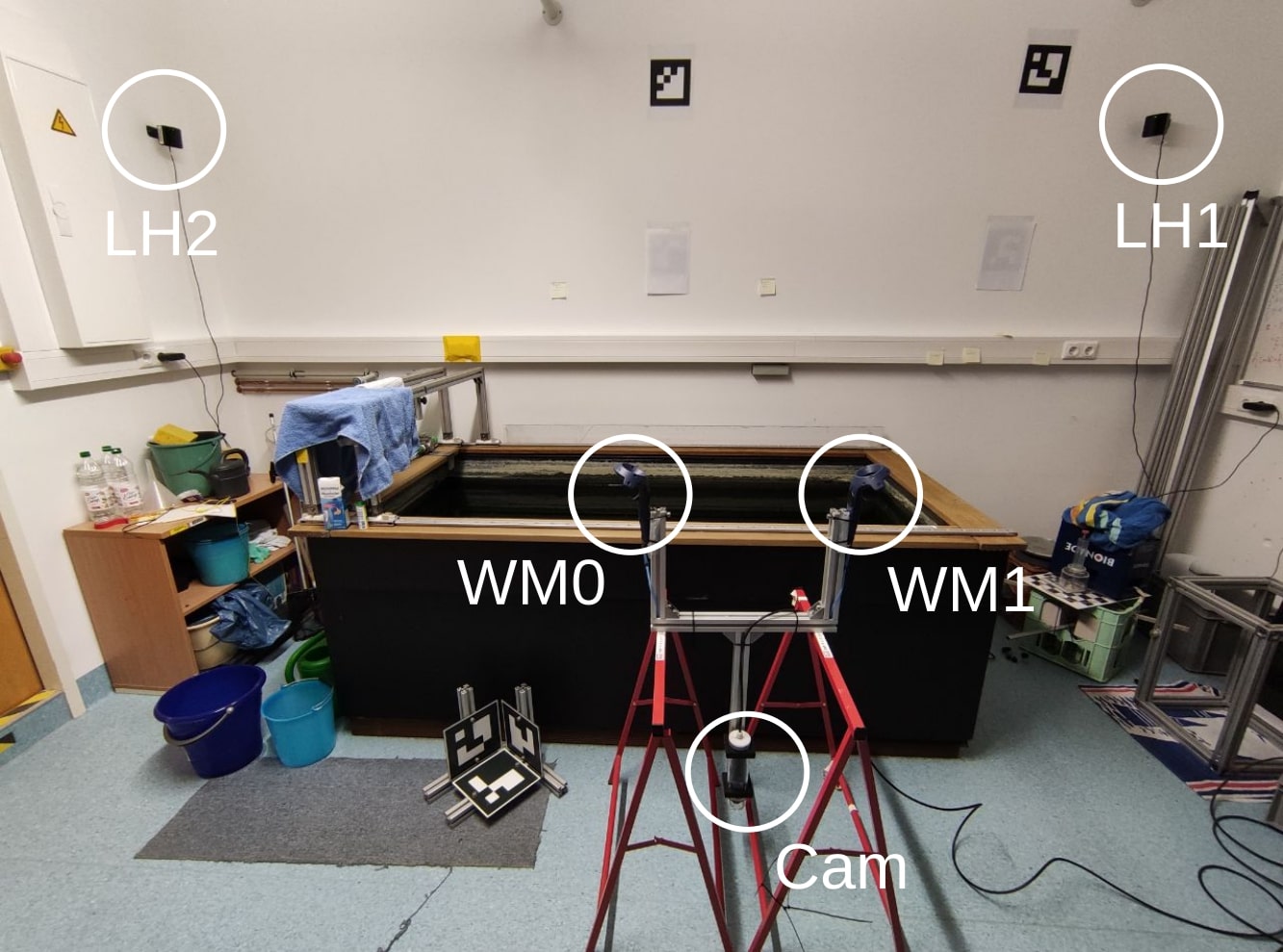}
        }
        \\
        \subfloat[Top view of simulation environment] {
            \includegraphics[height=0.35\textwidth]{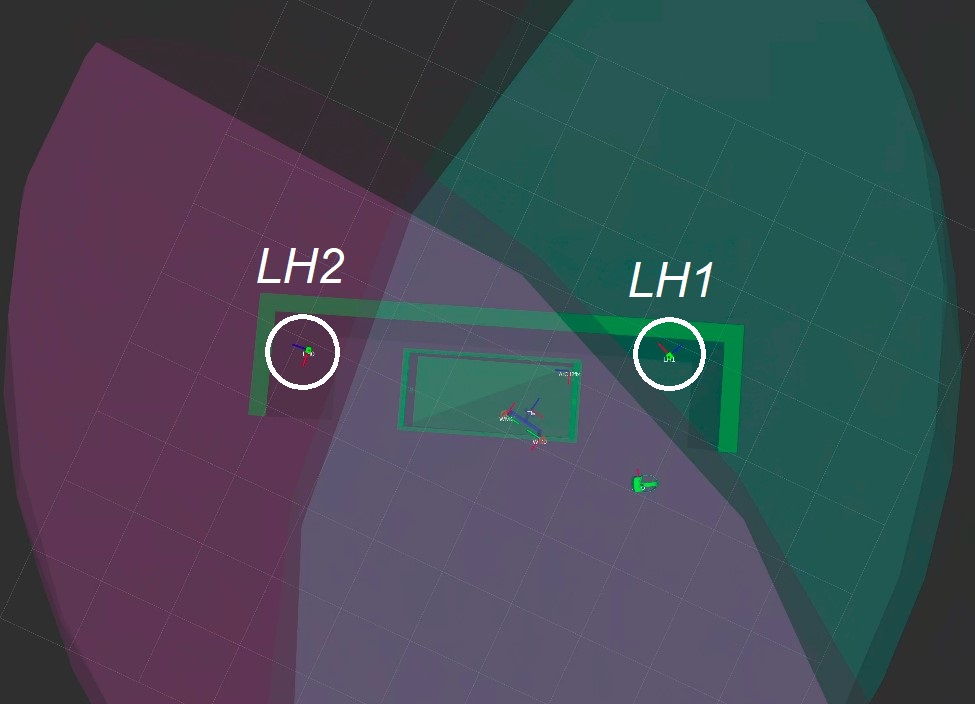}
        }
    \end{center}
    \caption{(a) System overview around the tank. (b) The system in a simulation with the tracking area of the lighthouses.} \label{fig:roomOverview}
\end{figure}

The so-called lighthouses of the Vive system emit light sweeps, which are detected by the photo diodes of the controllers.
The horizontal and vertical angle to the lighthouse can be calculated from the measured time of arrival and thus the absolute pose can be determined in real time.
In addition, IMU measurements are merged into the pose calculation to improve the precision of relative movements.
The arrangement of the lighthouses is selected in such a way that the center of the measurement environment with the water tank is optimally located in the tracking area.
This also considers that the user of the system does not stand between the lighthouses field of view and the controllers during the measurement.
The lighthouses are mounted at a height of 2.3 meters above the floor, with about 4.2 meters between them.
In this setup, we did not observe any problems stemming from reflections from the water surface.
The surroundings of the measurement setup and the tracking area of the lighthouses with the cone-shaped coloring in magenta for the first lighthouse and cyan for the second lighthouse are shown in \autoref{fig:roomOverview}.

\subsection{Camera Stick}
\begin{figure}[!t]
    \begin{center}
        \subfloat[Camera stick in real environment]{
            \includegraphics[height=0.35\textwidth]{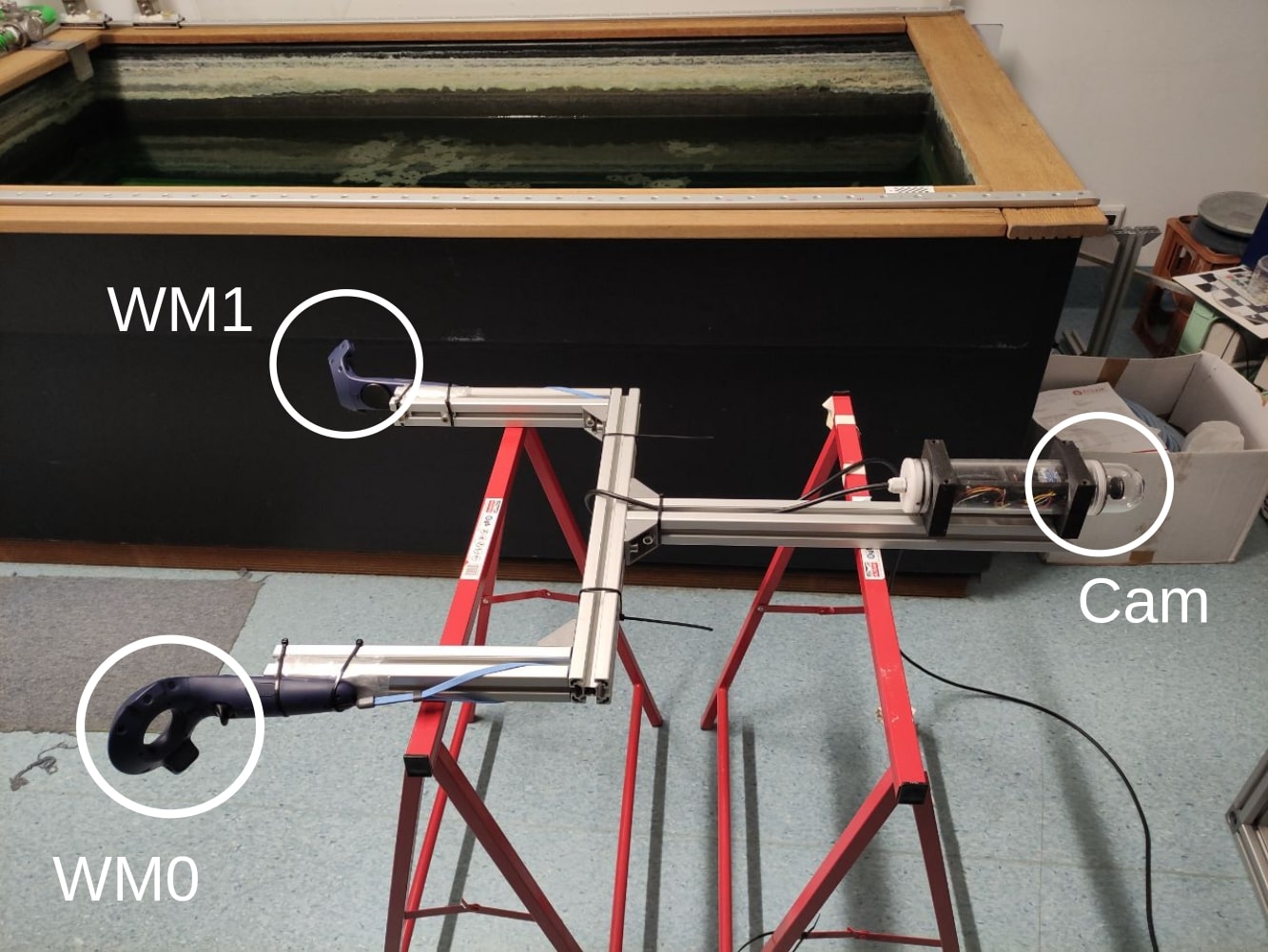}
        }
        \\
        \subfloat[Camera stick in virtual environment] {
            \includegraphics[height=0.35\textwidth]{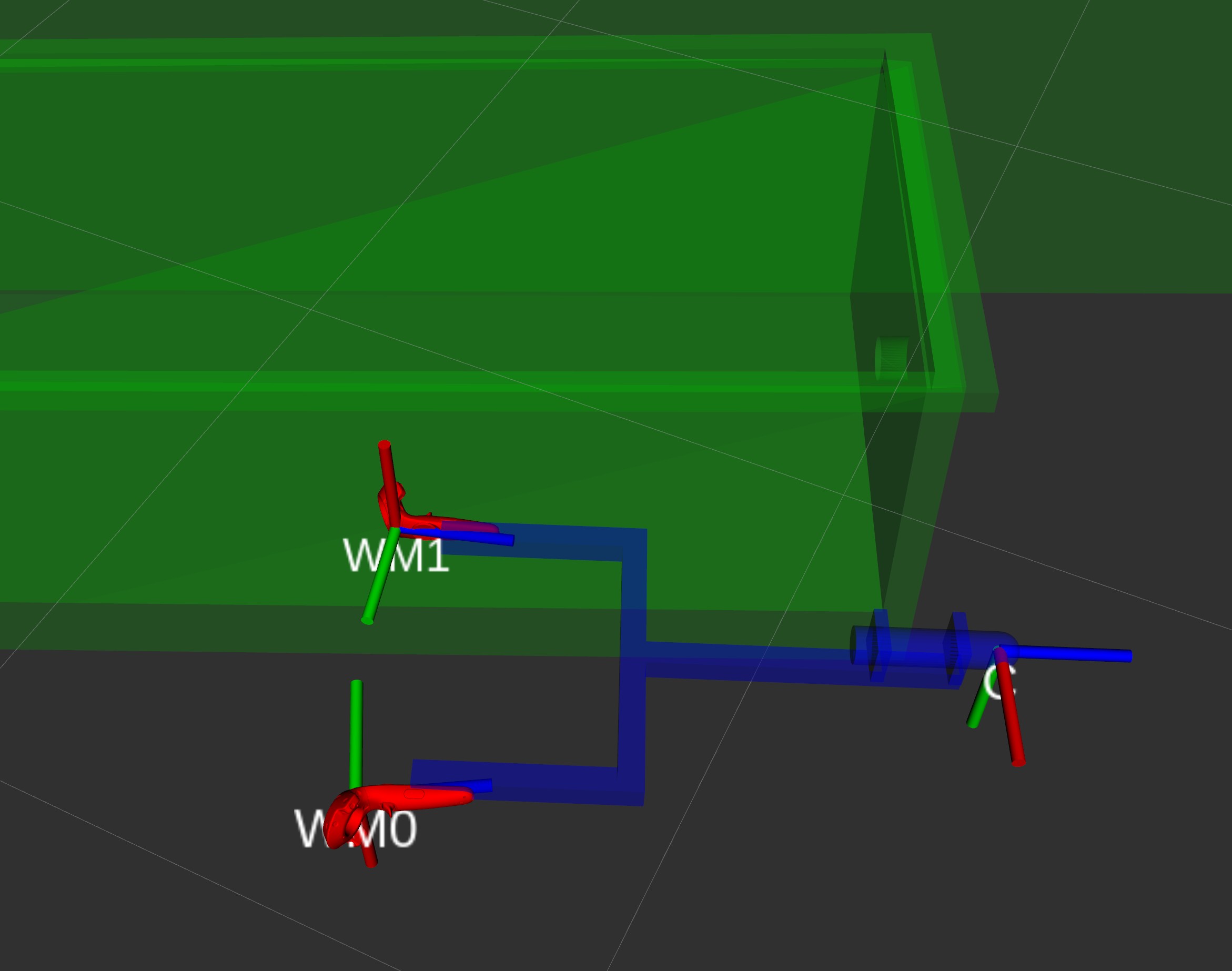}
        }
    \end{center}
    \caption{(a) Camera stick in the measurement environment. (b) Camera stick in the real time visualization using RViz.} \label{fig:camStickVisualization}
\end{figure}

The camera stick establishes the rigid mechanical connection between the camera (underwater) and the tracking system.
The camera's pose therefore relates to the controller's pose by a rigid transformation.
In \autoref{fig:camStickVisualization} the camera stick is shown with the camera located on the right side, so that it is about 0.9 m below the controllers.
Thus, this part can be kept submerged while the upper elements of the stick with the controllers remain above the water surface. 
The use of two controllers should make it possible to mutually play off their individual tracking errors and to achieve an optimized result. 
Empirical experiments were used to determine in which relative positioning the controllers exhibit the smallest tracking error.
This is achieved by rotating them 180° to each other, orienting them upright and having a distance between the origins of the rigid bodies of about 0.52 m.
As shown in \autoref{fig:camStickVisualization}, the real environment is visualized in real time in the ROS visualization (RViz).

\subsection{Underwater Camera}

\begin{figure}[!t]
    \begin{center}
        \subfloat[Camera top view]{
            \includegraphics[height=0.5\textwidth]{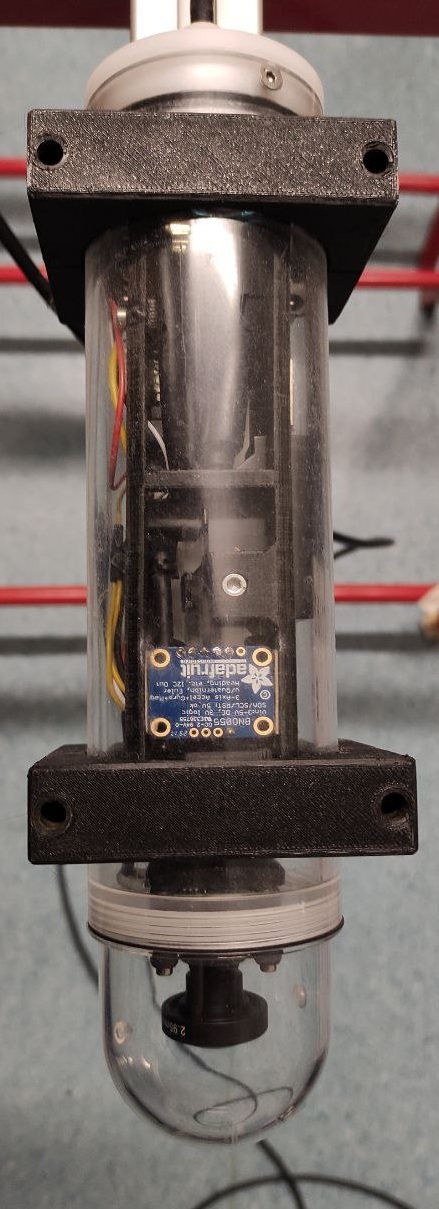}
        }
        \subfloat[Camera side view] {
            \includegraphics[height=0.5\textwidth]{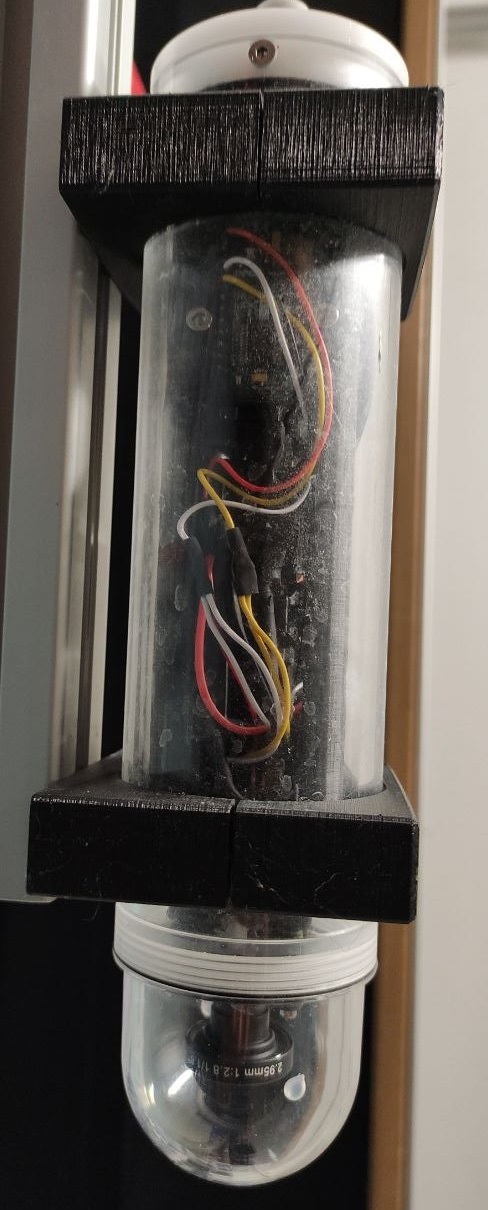}
        }
    \end{center}
    \caption{(a) The camera shown from the front. (b) The camera shown from the side.} \label{fig:cameraImage}
\end{figure}

The characteristics of the Basler Machine Vision camera used are listed in \autoref{tab:camLens}.
While \autoref{fig:cameraImage} shows the custom-build camera underwater housing in detail.
It is comprised of the camera in an adjustable mount, to enable setting its offset with respect to the dome port,
an Arduino to process the data of IMUs and an USB hub to combine all data into one cable, which leaves the pressure housing.

We then employ the following calibration scheme, to determine the extrinsics and intrinsics of the camera:
Initially, we estimate the cameras distortion parameters in air based on a fisheye model with a reprojection error of 0.22px. 
Subsequently, we use the approaches presented in \cite{she2019adjustment,she2022refractive}, to obtain and correct the camera's displacement with respect to the dome center.
This step ensures that refraction effects due to the traversal of the light rays through interfaces between media with different optical densities are omitted.
The Hand-Eye calibration (see \autoref{ssec:handeye}) can now be carried out with the centered camera in air. 
For the actual underwater application, we again used a fisheye model and reached a reprojection error of 0.55px on a calibration set taken in a clear underwater setting.
The last step and the dome centering are carried out to eliminate all refraction-based effects such that the underwater sets can be used for pure radiometric problems.

\begin{table}[!t]
	\caption{Properties of the Basler camera \citep{baslerCam} and lens \citep{baslerLens} used.}\label{tab:camLens}
	\begin{center}
		\begin{tabular}{|l|p{3cm}|}
			\hline
			Property camera & Value \\
			\hline
			\hline
			Max. Frame Rate & 60 FPS \\
			\hline
			Resolution & 2 MP \\
			\hline
			Color & Color and gray scale \\
			\hline
			Resolution (HxV) & 1600 px x 1200 px \\
			\hline
			Pixel Size (H x V) & 4.5 µm x 4.5 µm \\
			\hline
			Pixel Bit Depth & 8, 12 bits \\
			\hline		
			Shutter & Global Shutter \\
			\hline
			\noalign{\vskip 5mm}
			\hline
			Property lens & Value \\
			\hline
			\hline
			Focal Length & 2,95 mm \\
			\hline
			Min. Working Distance & 300 mm \\
			\hline
			Resolution & 5 MP \\
			\hline
			Angle of View (D / H / V) & 180° / 143° / 106° \\
			\hline
		\end{tabular}
	\end{center}	
\end{table}

\subsection{Calibration Target}

\begin{figure}[!t]
    \begin{center}
        \subfloat[Calibration target]{
            \includegraphics[height=0.16\textwidth]{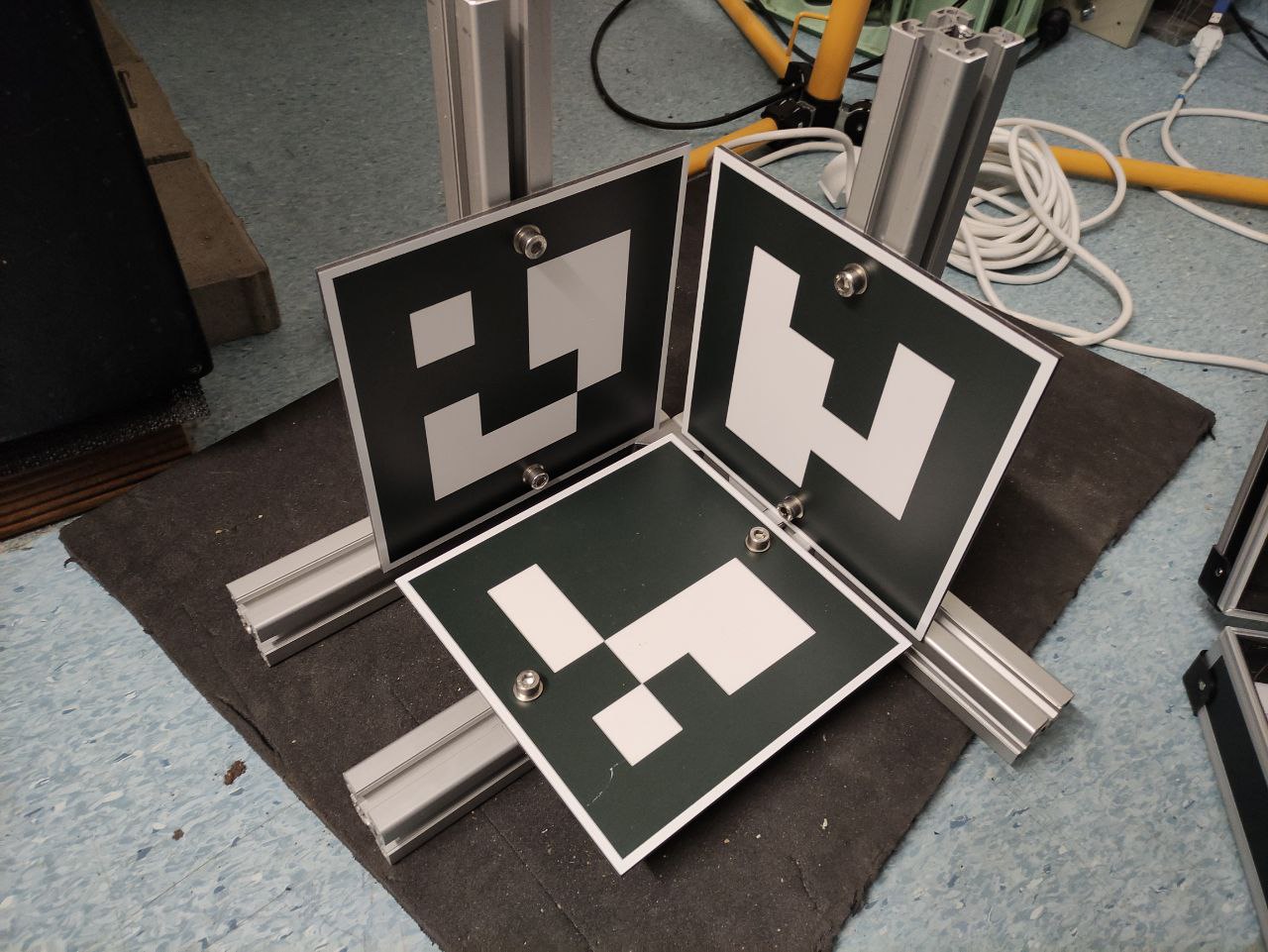}
        }        
        \subfloat[Calibration target detected] {
            \includegraphics[height=0.16\textwidth]{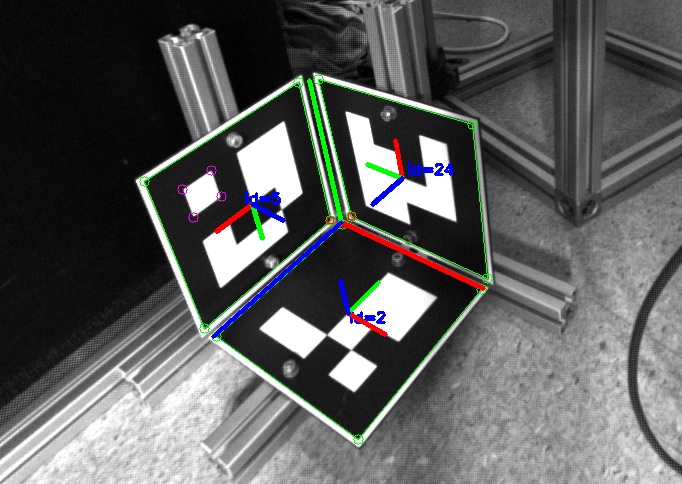}
        }
    \end{center}
    \caption{(a) The optical target used for the Hand-Eye calibration. (b) Detection of the calibration target.} \label{fig:calibTarget}
\end{figure}

A calibration target is required for the system calibration and verification.
This serves as a fixed point in 3D space and is composed of three ArUco markers, which can be detected by the camera.
The pose detection during image processing is improved by calculating the transformations to a joint board pose from several markers.
Initially, the relative poses of the three markers have been optimized in a least square scheme using the Ceres solver \citep{ceres}.
The three markers are assembled in the form of the inside view of a 3D angle, which is shown in \autoref{fig:calibTarget}.

\subsection{Motion capture}

The HTC Vive Pro system is a room scale system with six degrees of freedom.
These are defined by the rotation around three axes and the position of objects in 3D space.
The measurement principle is outside-in tracking, in which the optical part, i.e. the lighthouses, is placed in fixed spatial location.
Each lighthouse covers a pyramidal volume of the tracking area with an opening angle of 150° in horizontal and 110° in the vertical direction.

For tracking, the lighthouses emit an infrared light flash over the entire tracking area, which is alternately followed by a horizontal and a vertical infrared light plane to scan the room.
This sequence is repeated periodically.
The horizontal and vertical planes of light are generated by a rotor, which rotates at 50 to 55 rotations per second, depending on the channel used.	
Since the objects in the tracking area require additional information from the lighthouses, an optical transmission method is used to send the Omnidirectional Optical Transmitter (OOTX) data continuously over an infrared channel from each lighthouse.
This data is used to uniquely identify the lighthouse and transmit its calibration data to the receivers with the global light flash.

The headset and the controllers of the HTC Vive have infrared sensors that measure the time between the arrival of the global light flash and the next sweeping plane.
The angle between the normal vector of the lighthouse and an infrared sensor can be calculated from the time differences for horizontal and vertical measurements, since the rotational speed of the rotor in the lighthouse is known \citep{nasaVive}.
Here, the lighthouses can be viewed as cameras, which determine 2D positions in the image with the infrared scanners.
This results in the horizontal and vertical direction in which a tracking object is located, starting from the origin of the lighthouse coordinate system.

Since the 3D positions of the infrared sensors on the headset and the controllers are known from the construction, 2D-3D correspondences result from their positions and the results of the lighthouse measurement \citep{viveResearch}.
By solving them, the absolute pose determination for headset and controllers can be accomplished.
In order to improve the detection of relative movements and to increase the frequency of pose calculation, the relative movement of the tracking devices are determined by a built-in IMU.
The inertial data includes linear acceleration and angular velocity, which are integrated into the absolute pose.
The deviations caused by the drift in the relative measurements are compensated by the lighthouse measurements.

\subsection{Hand-Eye calibration}
\label{ssec:handeye}

\begin{figure}[!t]
    \begin{center}
        \includegraphics[width=.5\textwidth]{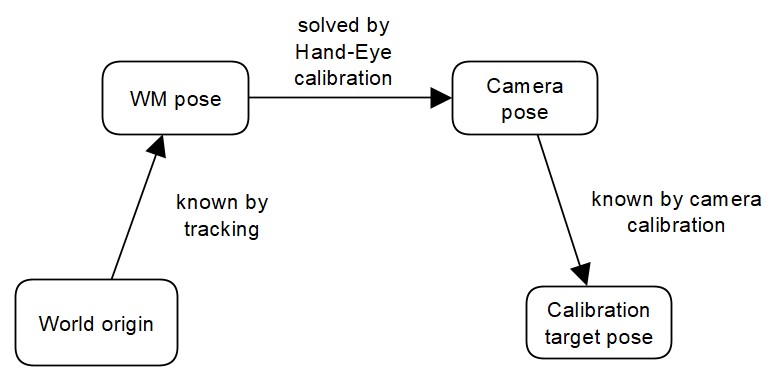}
    \end{center}
    \caption{Overview of the requirements for Hand-Eye calibration. \label{fig:handeyeOverview} }
\end{figure}

The term Hand-Eye calibration is derived from the field of robotics where the camera is called the eye and the joint with its gripper is called the hand.
By using a controller as a hand, its pose in the world coordinates is given by the VR tracking system.
The calibration deals with the calculation of the relative transformation $\mathbf{TF}_{WM2Cam}$ from a controller to the camera pose \citep{handeyeBasics}.	
For the calculation, the transformation $\mathbf{TF}_{World2WM_i}$ from the world origin to the controller in 3D space and the relative transformation $\mathbf{TF}_{Cam_i2Target}$ between the camera and the calibration target are needed at the same time \citep{tsai}.
The transformation from the camera to the calibration target results from the image processing of the captured images with the known intrinsic camera parameters.
The connections between the transformations for the Hand-Eye calibration are shown in \autoref{fig:handeyeOverview}.

\begin{figure}[!t]
    \begin{center}
        \includegraphics[width=.5\textwidth]{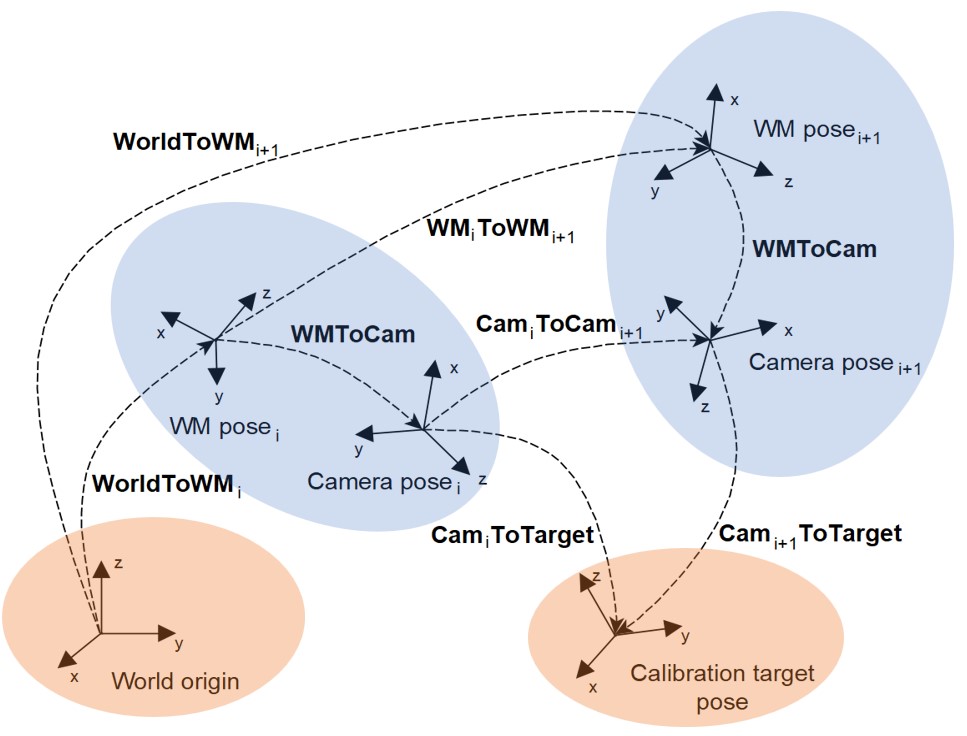}	
    \end{center}
    \caption{Overview of the transformations for two consecutive data sets of the Hand-Eye calibration. \label{fig:handeyeSetOverview} }
\end{figure}

With a known movement, a system of equations can be set up, and with enough samples, an accurate calculation of the calibration can be achieved.
The movement must have sufficient degrees of freedom so that the three-dimensional relationship between the camera and the controllers can be fully determined.
To do this, it is necessary to rotate the camera around at least two axes while keeping the focus on a calibration target which is placed in the world system and will not be moved during calibration \citep{tsai}.
The set also contains the transformations $\mathbf{TF}_{WM_i2WM_{i+1}}$ between two consecutive controller poses and $\mathbf{TF}_{Cam_i2Cam_{i+1}}$ between two consecutive camera poses.
The result of the Hand-Eye calibration describes the relative transformation $\mathbf{TF}_{WM2Cam}$ between the origin of the controllers and the lens of the camera.
This is calculated across the entire data set and is identical for each data pair due to the rigid transformation between the controllers and camera.
\autoref{fig:handeyeSetOverview} explains the principle of calibration.

Since two controllers are used on the camera stick, an individual transformation to the camera pose is required for each controller.
These two transformations are determined using the same set of images.
Each sample is therefore composed of the two controller poses in the world system and the transformation from the camera to the calibration target.
To minimize the influence of tracking and image processing deviations on the calibration result, the set size is defined as 50 samples.
In addition, the calibration is performed 20 times, so that a set of 20 transformations is available for each transformation required.
The set size and number of sets were determined empirically.
Subsequently the result is optimized by linearizing the non-linear pose-optimization problem and solving it with a least squares approach with the Ceres solver \citep{ceres}.
Since these transformations each correspond to the result of an entire Hand-Eye calibration, the assumption is made that the calibration process has already compensated the outliers.
Consequently, all transformations for the Ceres optimization are defined with the same uncertainty.

\subsection{Unscented Kalman Filter}

The Kalman Filter (KF) describes a mathematical model for integrating and fusing successive, noisy measurements of a system in a linear environment with one or multiple sensors.
The KF is real-time capable and therefore suitable for tracking.
The goal is to merge the measurements from the two controllers so that the overall tracking error is minimized.
The first order Markov assumption states, that all prior information is aggregated in the current state.
Hence, only the previously calculated state, a motion model, and current input measurement, as well as their uncertainties are considered for the estimation of the next state.
For a detailed description of the KF, refer to \cite{kf}.

In this work, the Unscented Kalman Filter (UKF) is used \citep{julierNewExtensionKalman1997,wanUnscentedKalmanFilter2000}.
The UKF has the advantage that it can be used in nonlinear systems.
In addition, the calculation is a second-order approximation, so that the use of Jacobian matrices is not necessary to set up the target covariance.
The UKF is based on statistical techniques and takes deterministic samples of a system.
A minimal number of samples around the mean, called sigma points, are used.
Using Newton's laws of motion, the unscented transform is applied to the set of sigma points by the motion model based on the observed velocity, angular velocity, and linear acceleration.
Then the new mean and its covariance are calculated by comparing the prediction with the next measurement.
A description of the UKF calculations can be found in \cite{ukf}.

A 15-dimensional state vector $\mathbf{x_{k}}$ is used for the UKF, which according to \autoref{eq:ukfStateVec} contains the position $p$, rotation $q$, velocity $v$, angular velocity $av$ and linear acceleration $a$
\begin{multline} \label{eq:ukfStateVec}
    \mathbf{x_{k}} = [p_x, p_y, p_z, q_x, q_y, q_z, \\ v_x, v_y, v_z, av_x, av_y, av_z, a_x, a_y, a_z]^T.
\end{multline}
The rotation is represented by a quaternion and only the imaginary components are considered.
To learn more about representing rotation in 3D space, we refer to \cite{representingAttitude}.
The real part can be calculated from the imaginary components after the UKF iteration.
The state vector $\mathbf{x_{k}}$ describes the camera pose and its motion in the world coordinate system.

In order to use the UKF camera pose as a later ground truth reference for the captured images, the poses must be synchronized with the image time stamps.
The ROS time is used as a common time basis.
This is time-synchronous for all components in the system.
However, the camera used does not allow the time stamp to be set directly for the image recording.
Consequently, this is set within the pylon software on the PC.
The image time stamps are defined in the middle of the exposure time.
Then the offset of the UKF camera pose to the recorded image is determined.
To do this, rotating movements are carried out over the calibration target, as well as forwards and backwards movements in the direction of the calibration target.
The evaluation of both experiments shows that the error due to the time difference is smaller than the error due to tracking and image processing.
Thus, the time-synchronous behavior between poses and image time stamps is given.

\section{Evaluation}
\label{sec:evaluation}

In the evaluation, the various system components are analyzed independently, and the precision of the overall system is determined.
Wherever possible, the respective experiment is reproduced with synthetic poses.
This procedure offers the advantage that the ground truth is always known for the synthetic poses.
A test scenario is used for the simulation, in which the synthetically generated camera stick is moved along the edges of a square with an edge length of about 1 m.
In doing so, rotations are performed on all coordinate axes.
Zero mean normal distributed noise is added to synthetically generated poses. Its standard deviation is based on the uncertainties observed in the real system.
Of course, least-squares optimization procedures will be able to find optimal solutions on such systematic noise patterns, which is a trivial insight.
However, synthetic ablation studies are the best we can do to justify each layer of complexity in the absence of ground truth data.
In top of that, the noise is applied on the input side of the evaluated modules, while the optimization happens on the output side.

The real system is analyzed by using the camera stick under dry conditions.
It is the goal to validate the system independently of the sources of interference that arise in an underwater application.
These can include, for example, blurred images and poor lighting conditions.
In order to represent the later application realistically, movements of the camera stick are carried out in the empty tank.

The evaluation procedure is based on the transformation representation in \autoref{fig:surviveWorld}.
These transformations are each characterized by uncertainties, which are to be analyzed one after the other.
For the sake of simplicity and only for the statistical analysis, independence between rotation and translation parts of the investigated transformations is assumed.
First, the tracking of the controllers is examined and then the Hand-Eye calibration is analyzed.
Finally, the precision of the UKF and the precision of the overall system are determined.

\subsection{Analysis of VR Tracking System}

This section defines the basis for an approximate determination of the precision of the tracking of the controllers.
The fundamental problem is that the precision cannot be directly determined from the tracking data, since the ground truth poses of the controllers are unknown.
The transformation $\mathbf{TF}_{WM}$ between the two controllers offers an approach for the analysis.
This transformation can be regarded as constant due to the rigid construction of the stick.
During tracking, the deviation of this transformation from the expected mean transformation can be determined.
This describes the combined tracking error of both controllers.

\begin{figure}[!t]
    \begin{center}
        \fbox{\includegraphics[width=.45\textwidth]{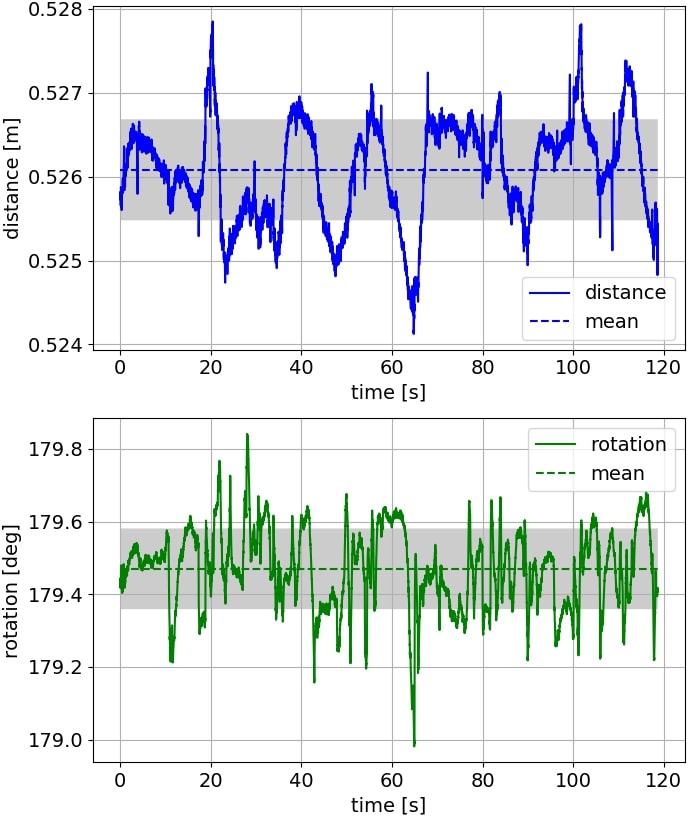}}
    \end{center}
    \caption{Deviation between the controller poses for distance and rotation over time for dataset 1 in \autoref{tab:realTrackingAnalysis}. The gray area represents the 1-$\sigma$-standard deviation. \label{fig:realTrackingAnalysis} }
\end{figure}

The mean values of the transformations between the controllers $\overline{TF}_{WM}$ over the datasets, their maximum deviation $\epsilon_{max}$ and the standard deviation $\sigma$ are analyzed.
\autoref{tab:realTrackingAnalysis} shows five datasets, each lasting 110-140 s.
This shows that the rigid transformation between the two controllers $\mathbf{TF}_{WM}$ is stable up to a few millimeters and fractions of a degree.
The average standard deviation $\overline{\sigma}$ of the transformation over the data sets is 1.12 mm and 0.12°.
Based on these values, the simulation of the controller poses is parameterized.
For dataset 1 in \autoref{tab:realTrackingAnalysis}, \autoref{fig:realTrackingAnalysis} shows the course of the absolute distance and rotation between the controllers over time.
For the simulation, the experiment is repeated and deviations of the transformation $\overline{TF}_{WM}$ are observed in the same order of magnitude.
An average maximum deviation $\overline{\epsilon}_{max}$ of 4.99 mm and 0.5° as well as a mean standard deviation $\overline{\sigma}$ of 1.29 mm and 0.13° result over several simulated data sets.

\begin{table}[!t]
    \caption{Tracking analysis for the real system based on the transformation between the two controllers.}\label{tab:realTrackingAnalysis}
    \begin{center}
        \begin{tabular}{|r|r|r|r|}
            \hline
            \multicolumn{1}{|c|}{Dataset} & \multicolumn{1}{|c|}{$\overline{TF}_{WM}$ [mm]} & \multicolumn{1}{|c|}{$\epsilon_{max}$ [mm]} & \multicolumn{1}{|c|}{$\sigma$ [mm]} \\
            \hline
            \hline		
            1 & 526.08 & 2.51 & 0.60 \\
            \hline		
            2 & 525.88 & 3.76 & 1.33 \\
            \hline		
            3 & 525.89 & 4.12 & 1.37 \\
            \hline			
            4 & 525.91 & 3.86 & 0.83 \\
            \hline		
            5 & 525.37 & 4.42 & 1.49 \\
            \hline				
            \noalign{\vskip 5mm}
            \hline
            \multicolumn{1}{|c|}{Dataset} & \multicolumn{1}{|c|}{$\overline{TF}_{WM}$ [°]} & \multicolumn{1}{|c|}{$\epsilon_{max}$ [°]} & \multicolumn{1}{|c|}{$\sigma$ [°]} \\
            \hline
            \hline		
            1 & 179.47 & 0.49 & 0.11 \\
            \hline	
            2 & 179.48 & 0.43 & 0.11 \\
            \hline		
            3 & 179.47 & 0.61 & 0.14 \\
            \hline			
            4 & 179.50 & 0.45 & 0.12 \\
            \hline		
            5 & 179.44 & 0.53 & 0.14 \\
            \hline	
        \end{tabular}
    \end{center}	
\end{table}

\subsection{Analysis of Hand-Eye Calibration}

In order to maintain the high precision of the tracking, an exact transformation from the controllers to the camera must be determined during the Hand-Eye calibration.
Therefore, in \autoref{tab:simulationHandEyeResults}, the effect of the optimization is first shown using the simulation.
Zero mean normal distributed noise with standard deviations based on the preceding analysis (1 mm and 0.1°) is added to each coordinate axis during simulation to the poses of the controllers.
To account for the imperfection of the pose estimation process from images of visual markers, random noise with the same parameters (zero mean, std. deviations 1 mm and 0.1°) is added to camera poses computed from images of markers.
For each data set, the average rigid transformation between the camera poses $\overline\epsilon_{TF}$ of the 20 calibrations is compared with the result of the optimization $\epsilon_{TF_{op}}$.
In addition, the average deviation of the individual camera poses from the ground truth pose $\overline\epsilon_{C}$ is compared with the optimized camera poses $\epsilon_{C1_{op}}$.
It is shown that the accuracy of the Hand-Eye calibration result is improved by a factor of approx. 4 on average by the optimization.

\begin{table}[!t]
    \caption{Hand-Eye calibration analysis based on simulated data.}\label{tab:simulationHandEyeResults}
    \begin{center}
        \begin{tabular}{|r|r|r|r|r|r|r|}
            \hline
            \multicolumn{1}{|c|}{\makecell{Data- \\ set}} & \multicolumn{1}{|c|}{\makecell{$\overline\epsilon_{TF}$ \\ \text{[mm]}}} & \multicolumn{1}{|c|}{\makecell{$\epsilon_{TF_{op}}$ \\ \text{[mm]}}} & \multicolumn{1}{|c|}{\makecell{$\overline\epsilon_{C1}$ \\ \text{[mm]}}} & \multicolumn{1}{|c|}{\makecell{$\epsilon_{C1_{op}}$ \\ \text{[mm]}}} & \multicolumn{1}{|c|}{\makecell{$\overline\epsilon_{C2}$ \\ \text{[mm]}}} & \multicolumn{1}{|c|}{\makecell{$\epsilon_{C2_{op}}$ \\ \text{[mm]}}} \\
            \hline
            \hline			
            1 & 1.88 & 0.36 & 2.23 & 0.25 & 2.33 & 0.14 \\
            \hline			
            2 & 2.03 & 0.46 & 2.07 & 0.35 & 2.18 & 0.70 \\
            \hline			
            3 & 2.09 & 0.26 & 1.65 & 0.57 & 1.86 & 0.33 \\
            \hline			
            4 & 2.20 & 0.62 & 1.82 & 0.73 & 2.05 & 0.67 \\
            \hline			
            5 & 1.92 & 0.26 & 2.30 & 0.61 & 1.93 & 0.76 \\
            \hline
            \noalign{\vskip 5mm}
            \hline
            \multicolumn{1}{|c|}{\makecell{Data- \\ set}} & \multicolumn{1}{|c|}{\makecell{$\overline\epsilon_{TF}$ \\ \text{[°]}}} & \multicolumn{1}{|c|}{\makecell{$\epsilon_{TF_{op}}$ \\ \text{[°]}}} & \multicolumn{1}{|c|}{\makecell{$\overline\epsilon_{C1}$ \\ \text{[°]}}} & \multicolumn{1}{|c|}{\makecell{$\epsilon_{C1_{op}}$ \\ \text{[°]}}} & \multicolumn{1}{|c|}{\makecell{$\overline\epsilon_{C2}$ \\ \text{[°]}}} & \multicolumn{1}{|c|}{\makecell{$\epsilon_{C2_{op}}$ \\ \text{[°]}}} \\
            \hline
            \hline		
            1 & 0.11 & 0.03 & 0.11 & 0.03 & 0.11 & 0.01 \\
            \hline			
            2 & 0.10 & 0.01 & 0.10 & 0.02 & 0.12 & 0.02 \\
            \hline			
            3 & 0.12 & 0.04 & 0.11 & 0.02 & 0.10 & 0.04 \\
            \hline			
            4 & 0.09 & 0.03 & 0.09 & 0.04 & 0.10 & 0.03 \\
            \hline			
            5 & 0.10 & 0.03 & 0.10 & 0.03 & 0.10 & 0.03 \\
            \hline
        \end{tabular}
    \end{center}	
\end{table}

\begin{figure}[!t]
    \begin{center}
        \fbox{\includegraphics[width=.45\textwidth]{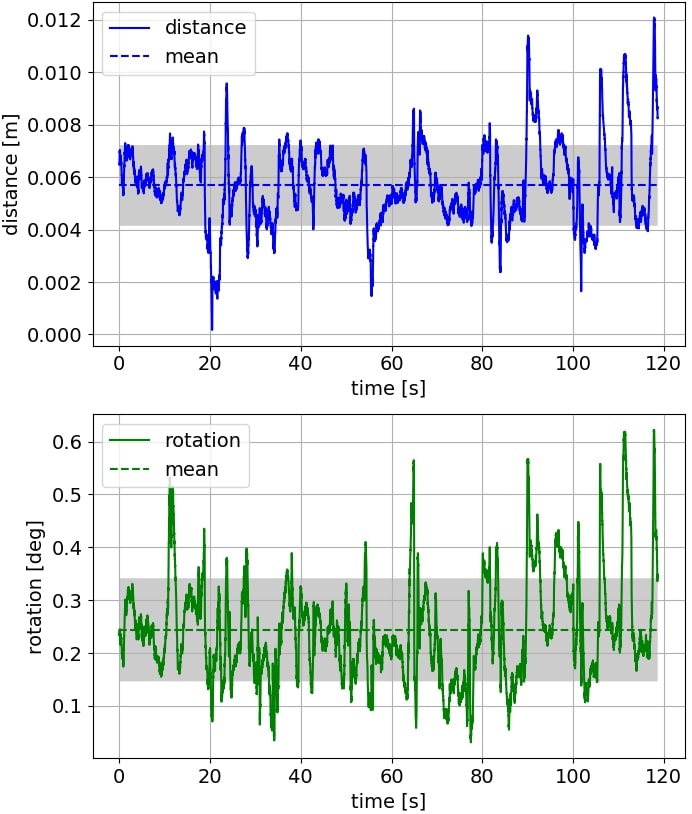}}
    \end{center}
    \caption{Deviation between the single camera poses for distance and rotation over time for dataset 1 in \autoref{tab:realHandEyeAnalysis}. The gray area represents the 1-$\sigma$-standard deviation. \label{fig:realHandEyeAnalysis} }
\end{figure}

Since the ground truth poses are unknown in the real system, this form of analysis cannot be repeated there.
Instead, the camera poses are analyzed after the Hand-Eye calibration.
For this purpose, the transformation between the two propagated camera poses $\mathbf{TF}_{Cams}$ is determined during the tracking.
This can then be used to determine the mean deviation $\overline{TF}_{Cams}$, as well as the maximum error $\epsilon_{max}$ and the standard deviation $\sigma$ of the transformation.
The errors shown in \autoref{tab:realHandEyeAnalysis} therefore contain both the calibration error and the error originating from the tracking system.
There is an average deviation between the camera poses $\overline{TF}_{Cams}$ of 6.57 mm and 0.29°.
For dataset 1 in \autoref{tab:realHandEyeAnalysis}, \autoref{fig:realHandEyeAnalysis} shows the course of the absolute distance and rotation between the single camera poses over time.
In the simulation, the experiment was repeated with several data sets.
The average deviation between the camera poses $\overline{TF}_{Cams}$ is 3.53 mm and 0.21°.
In the real system, the errors are bigger due to real effects such as the systematic drift and outliers in the tracking.
The position of the camera is particularly affected by small errors in rotation of the controllers.
This can easily be explained by the leverage effect stemming from the physical design of the camera stick (see \autoref{fig:camStickVisualization}).

\begin{table}[!t]
    \caption{Camera poses analysis for the real system based on the transformation between the Cameras.}\label{tab:realHandEyeAnalysis}
    \begin{center}
        \begin{tabular}{|r|r|r|r|}
            \hline
            \multicolumn{1}{|c|}{Dataset} & \multicolumn{1}{|c|}{$\overline{TF}_{Cams}$ [mm]} & \multicolumn{1}{|c|}{$\epsilon_{max}$ [mm]} & \multicolumn{1}{|c|}{$\sigma$ [mm]} \\
            \hline
            \hline		
            1 & 5.74 & 6.34 & 1.51 \\
            \hline		
            2 & 6.73 & 14.42 & 2.36 \\
            \hline		
            3 & 6.38 & 11.06 & 2.30 \\
            \hline			
            4 & 5.96 & 8.80 & 1.73 \\
            \hline		
            5 & 8.05 & 7.01 & 1.88 \\
            \hline				
            \noalign{\vskip 5mm}
            \hline
            \multicolumn{1}{|c|}{Dataset} & \multicolumn{1}{|c|}{$\overline{TF}_{Cams}$ [°]} & \multicolumn{1}{|c|}{$\epsilon_{max}$ [°]} & \multicolumn{1}{|c|}{$\sigma$ [°]} \\
            \hline
            \hline		
            1 & 0.25 & 0.38 & 0.09 \\
            \hline	
            2 & 0.29 & 0.73 & 0.10 \\
            \hline		
            3 & 0.30 & 0.50 & 0.12 \\
            \hline			
            4 & 0.26 & 0.59 & 0.08 \\
            \hline		
            5 & 0.37 & 0.36 & 0.10 \\
            \hline	
        \end{tabular}
    \end{center}	
\end{table}

\subsection{Analysis of Filtering (UKF)}

In order to determine the precision of the overall system, the calibration target is first measured in the world system according to \autoref{fig:surviveWorld} via a chain of transformations.
For this purpose, a large set of poses is recorded, and the optimal calibration target pose is then determined using a pose-optimization based on the Ceres library \cite{ceres}.
This makes it possible to compare an optically determined camera pose based on the visual markers with the camera pose computed by the proposed tracking system.
In addition to the errors from tracking and Hand-Eye calibration, errors from image processing and the time offset between the tracking poses and the image time stamps are also considered.
Due to the slow-motion speed during the later experiments, a small deviation in synchronization between image time stamp and camera pose time stamp will not lead to a large spatial deviation.
As an example, the average movement speed during the experiment is 56.8 mm/s and the average frame rate of the camera is 7.7 Hz.
A large synchronization error of ten percent of a time interval between two images is assumed for the calculation.
This would lead to a systematic position error of 0.74 mm between the true camera pose during image acquisition and the tracked camera pose from the external reference system.
However, such a large time offset would have been noticed during the calibration due to the systematic errors.
The remaining time error can therefore only be a few milliseconds and therefore the resulting position error is estimated at a maximum of 0.2 mm.
Consequently, time synchronicity is not a major problem for the planned application.

\begin{figure}[!t]
    \begin{center}
        \fbox{\includegraphics[width=.45\textwidth]{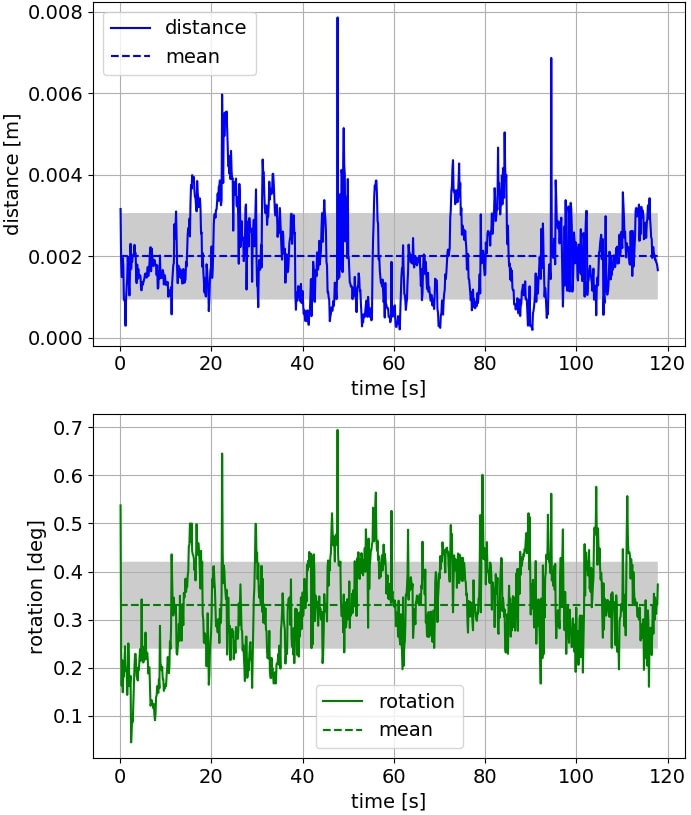}}
    \end{center}
    \caption{Deviation between the optical determined camera pose and the tracked camera pose after the UKF for distance and rotation over time for dataset 2 in \autoref{tab:realUkfAnalysis}. The gray area represents the 1-$\sigma$-standard deviation. \label{fig:realUkfAnalysis} }
\end{figure}

\autoref{tab:realUkfAnalysis} compares the mean deviations from the optically determined camera pose to the individual camera poses $\overline{\epsilon}_{C}$ and the UKF camera pose $\overline{\epsilon}_{C_{UKF}}$.
It is shown that the camera pose determined by the suggested system with the UKF is more precise than the individual camera poses using only one controller.
This results in a mean deviation of 2.51 mm and 0.26° from the UKF filtered camera pose to the optimized camera pose based on visual markers optimized across all datasets.
For dataset 2 in \autoref{tab:realUkfAnalysis}, \autoref{fig:realUkfAnalysis} shows the course of the absolute deviation for distance and rotation between the tracked camera pose and optical determined camera pose over time.
The experiment was repeated in the simulation using a perfect Hand-Eye calibration.
The comparison of the UKF camera pose to the ground truth camera pose results in an average deviation of 2.44 mm and 0.09° over several data sets.

\begin{table}[!t]
    \caption{Camera poses analysis for the real system compared to the optical camera pose.}\label{tab:realUkfAnalysis}
    \begin{center}
        \begin{tabular}{|r|r|r|r|}
            \hline
            \multicolumn{1}{|c|}{Dataset} & \multicolumn{1}{|c|}{$\overline{\epsilon}_{C_1}$ [mm]} & \multicolumn{1}{|c|}{$\overline{\epsilon}_{C_2}$ [mm]} & \multicolumn{1}{|c|}{$\overline{\epsilon}_{C_{UKF}}$ [mm]} \\
            \hline
            \hline		
            1 & 3.61 & 5.49 & 3.20 \\
            \hline	
            2 & 3.83 & 2.91 & 2.01 \\
            \hline		
            3 & 4.78 & 2.63 & 2.46 \\
            \hline			
            4 & 5.28 & 3.31 & 2.40 \\
            \hline		
            5 & 6.09 & 3.61 & 2.47 \\
            \hline				
            \noalign{\vskip 5mm}
            \hline
            \multicolumn{1}{|c|}{Dataset} & \multicolumn{1}{|c|}{$\overline{\epsilon}_{C_1}$ [°]} & \multicolumn{1}{|c|}{$\overline{\epsilon}_{C_2}$ [°]} & \multicolumn{1}{|c|}{$\overline{\epsilon}_{C_{UKF}}$ [°]} \\
            \hline
            \hline		
            1 & 0.32 & 0.38 & 0.32 \\
            \hline	
            2 & 0.30 & 0.39 & 0.33 \\
            \hline		
            3 & 0.23 & 0.32 & 0.22 \\
            \hline			
            4 & 0.25 & 0.30 & 0.21 \\
            \hline		
            5 & 0.25 & 0.34 & 0.23 \\
            \hline	
        \end{tabular}
    \end{center}	
\end{table}

\section{Discussion}
\label{sec:discussion}

\begin{figure*}
    \begin{center}
       \subfloat[Trajectory of the experiment, mimicking a lawn mower pattern typically executed by AUVs]{
           \includegraphics[width=.99\textwidth]{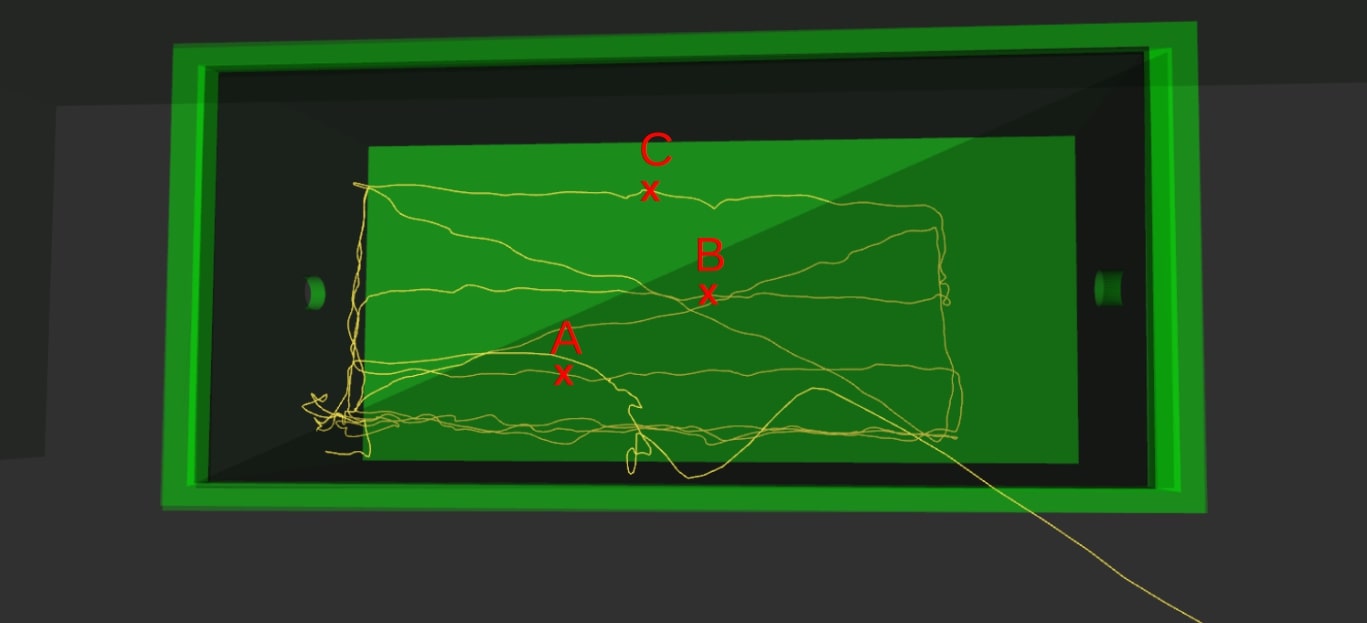}
       }
       \\~\\~\\
       \subfloat[Selected RAW images A, B and C from left to right] {
           \includegraphics[width=0.33\textwidth]{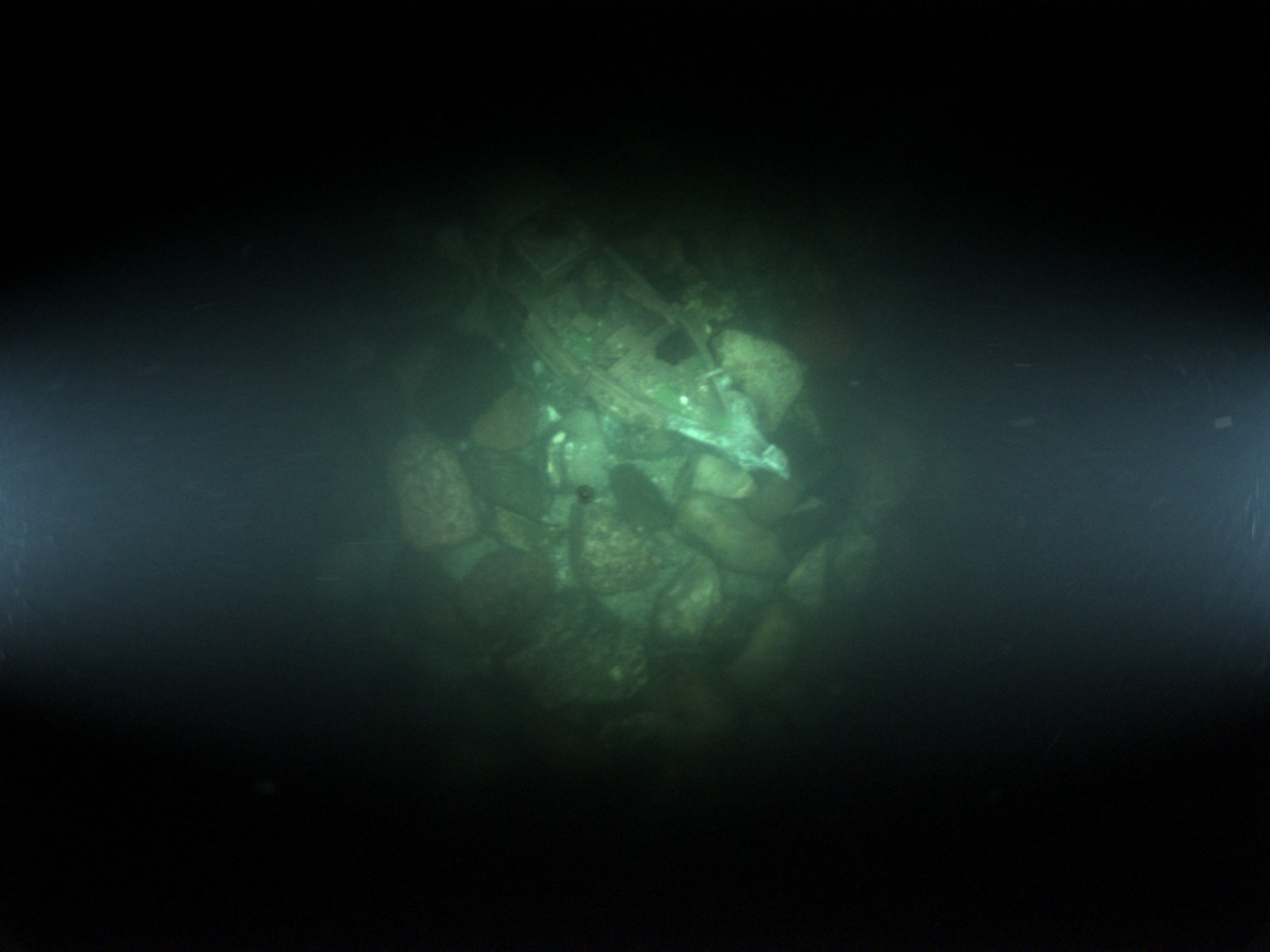}
           \includegraphics[width=0.33\textwidth]{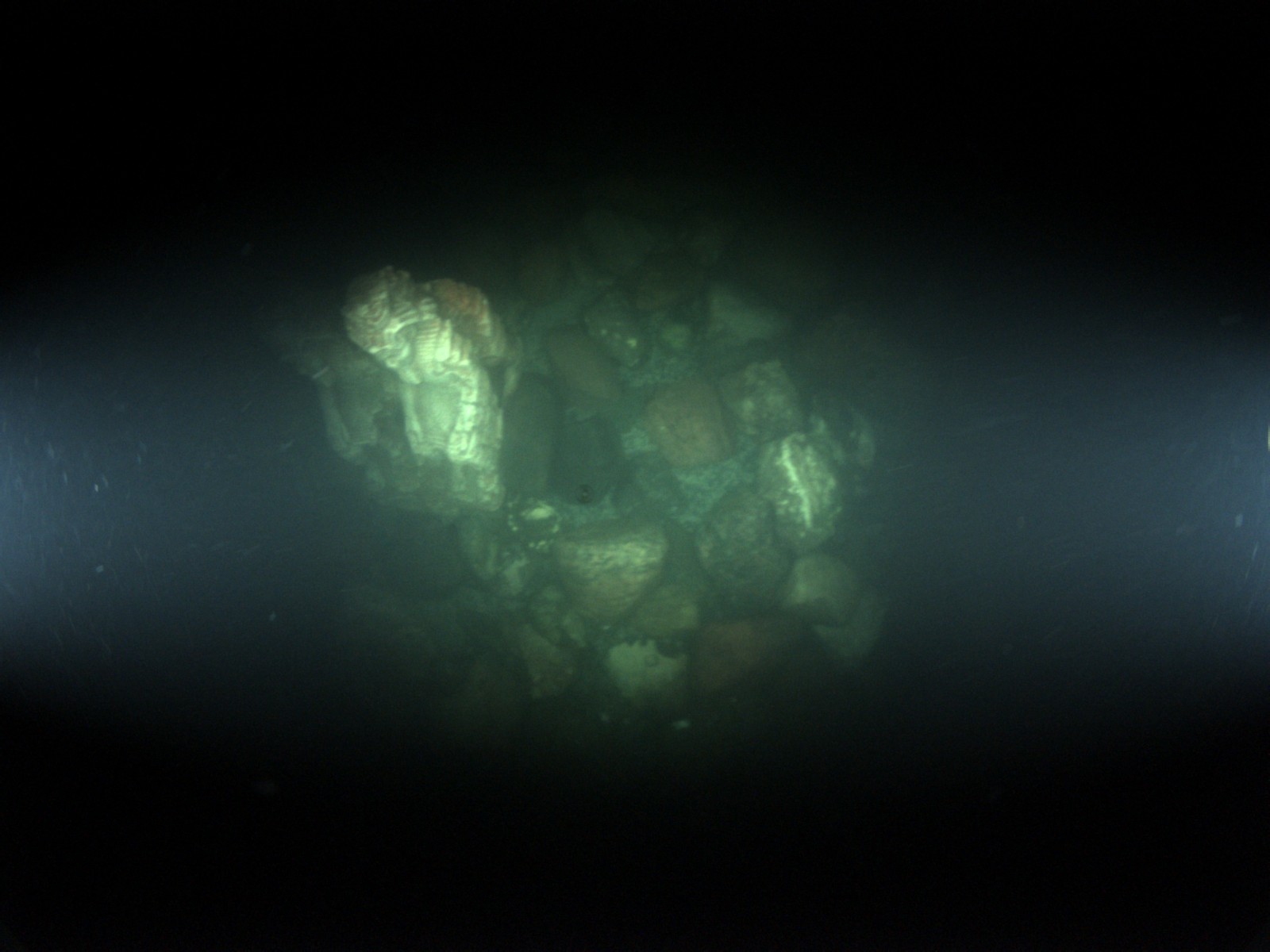}
           \includegraphics[width=0.33\textwidth]{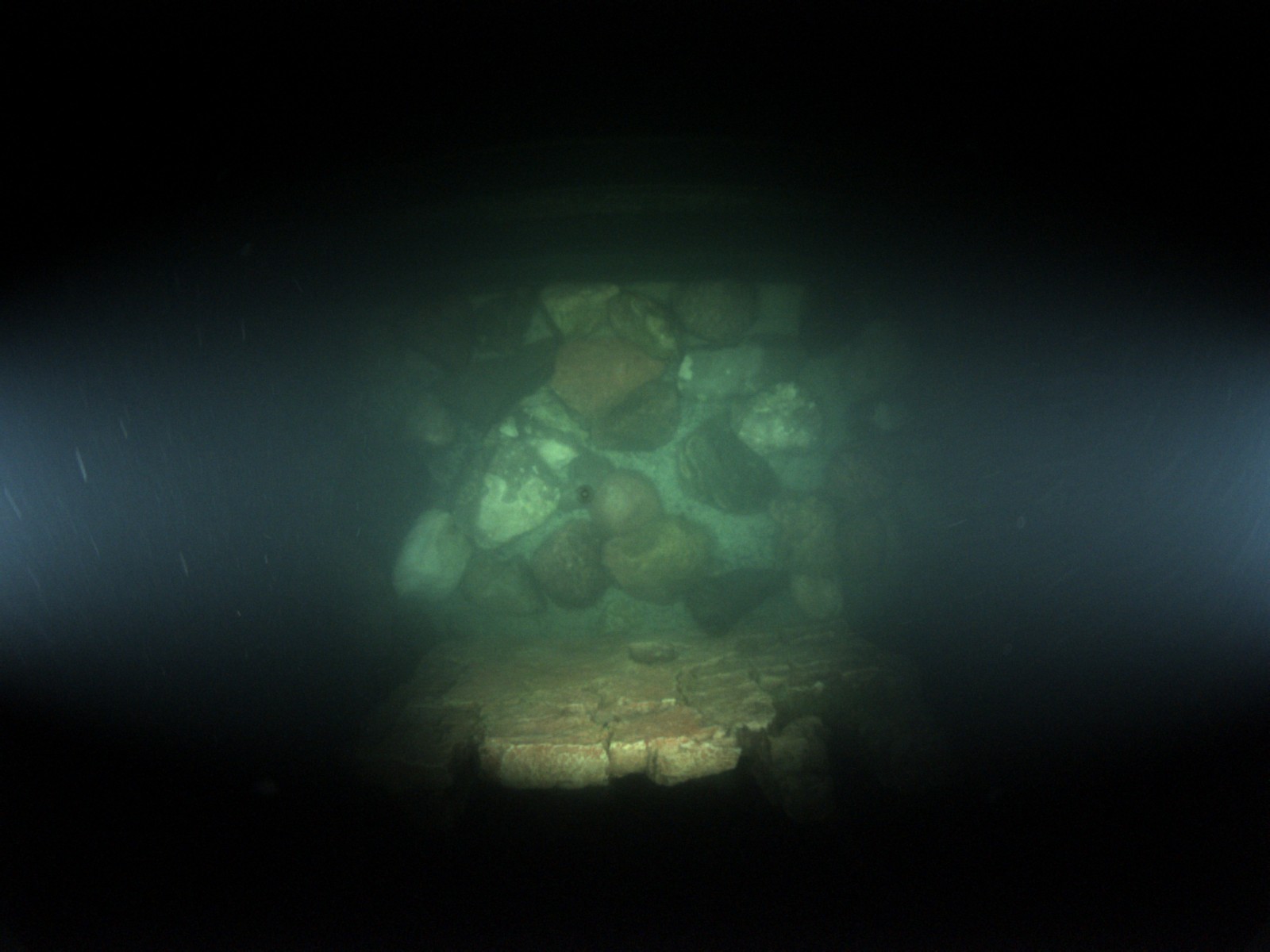}
       }
    \end{center}
    \caption{(a) Trajectory of the underwater experiment and three example positions of captured images. (b) The corresponding captured images during the experiment.} \label{fig:underwaterExperiment}
\end{figure*}

One difficulty in evaluating the system was that the ground truth poses of controllers and camera are not known in the real world.
The simulation uses synthetic poses to show that the developed algorithm performs well.
For this purpose, the system is realistically simulated in terms of tracking and error propagation.
Using this parameterization, it can be shown that the camera pose has an average deviation of 2.44 mm and 0.09° from ground truth in the simulation.

In the real system, a previously measured optical ground truth reference is used.
This system behaves very similar to the simulation in terms of tracking and error propagation, and the result for the camera pose is also in the same order of magnitude with an average deviation of 2.51 mm and 0.26°.
This proves that the approach developed for the external tracking system and the algorithms used are performing well.
Although this error will be scaled in the envisioned application, the tracking should still vastly outperform any pose estimation data captured in the deep sea.

Furthermore, it is shown that the transformation to the lower end of the camera stick only increases the error by a factor of 2-3 due to the leverage effect.
This is achieved by optimally fusing the resulting poses based on the two individual controllers using the UKF.
When replacing the commercially available drivers of the VR tracking system with open-source drivers (Libsurvive \citep{libsurvive}) the errors are significantly increased.
These increased errors could possibly be caused by imperfect parameters of the open-source algorithm.
However, the closed-source driver has limitations in terms of accessing raw data, the uncertainties, the algorithm and the precise timing.
Therefore, future work will deal with developing an optimizer for the open-source drivers in order to operate the external reference system with an precision of the same order of magnitude.

\begin{figure*}[!ht]
    \begin{center}
       \subfloat[In air]{
           \includegraphics[width=0.33\textwidth]{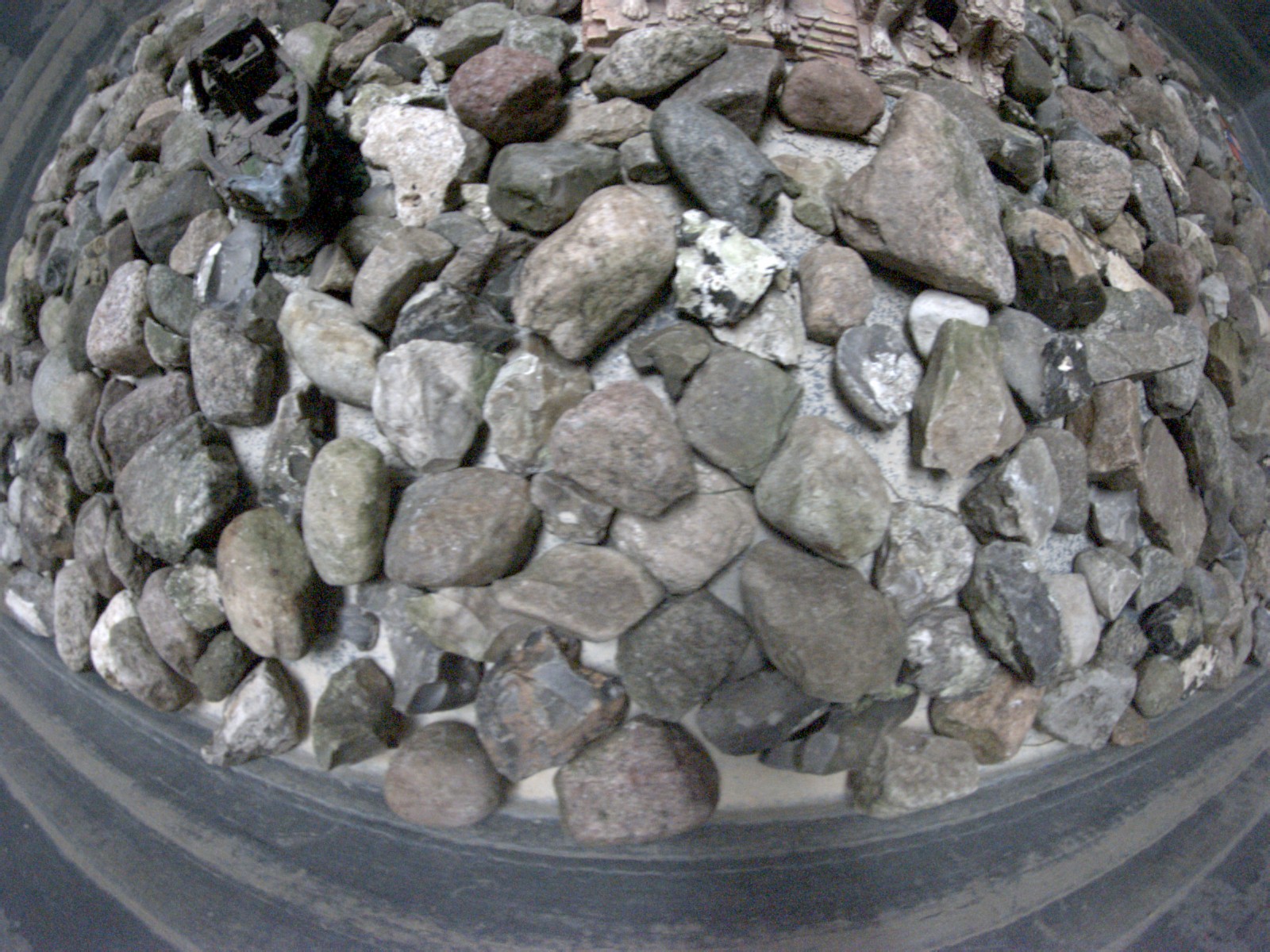}
           \includegraphics[width=0.33\textwidth]{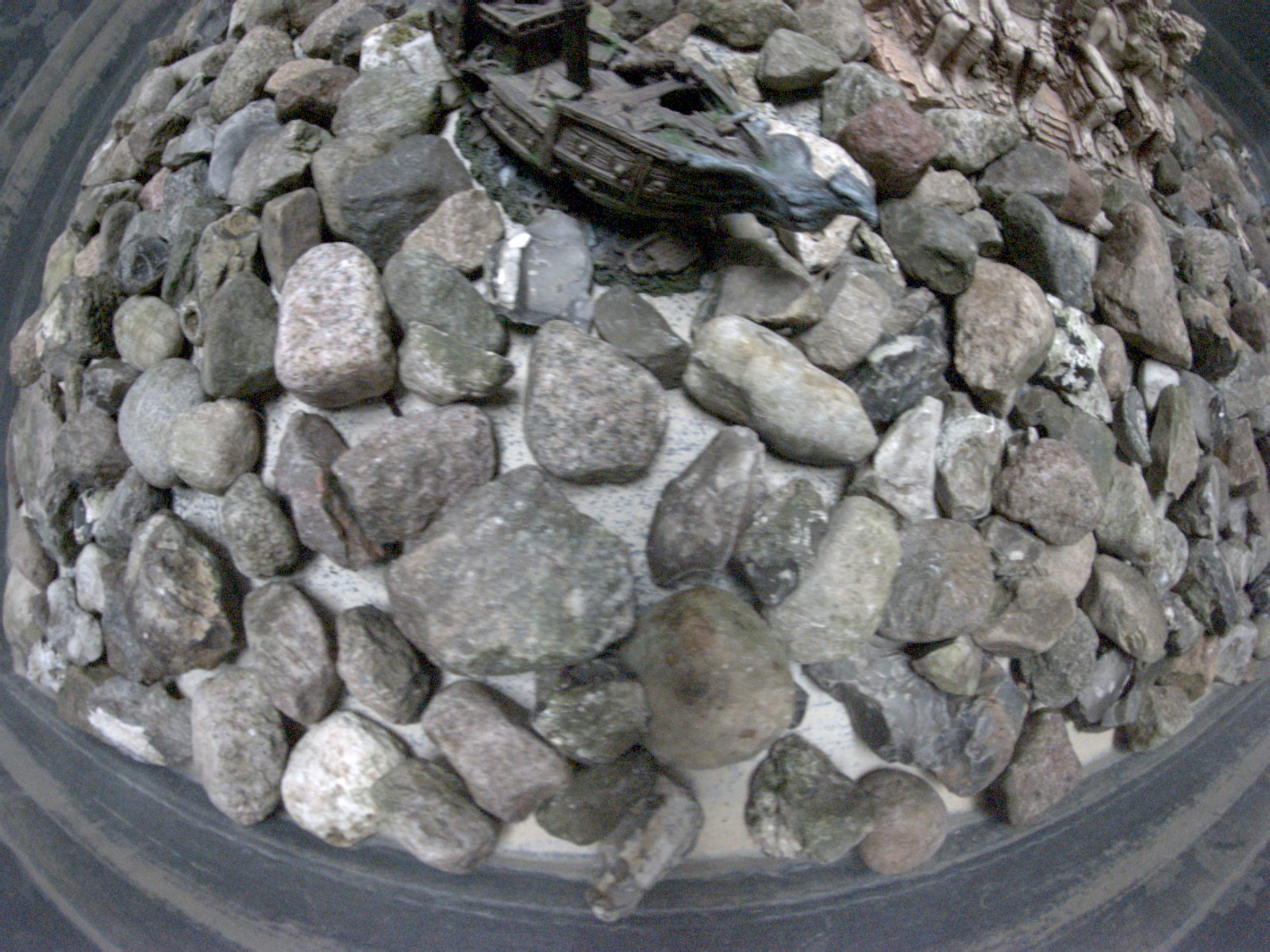}
           \includegraphics[width=0.33\textwidth]{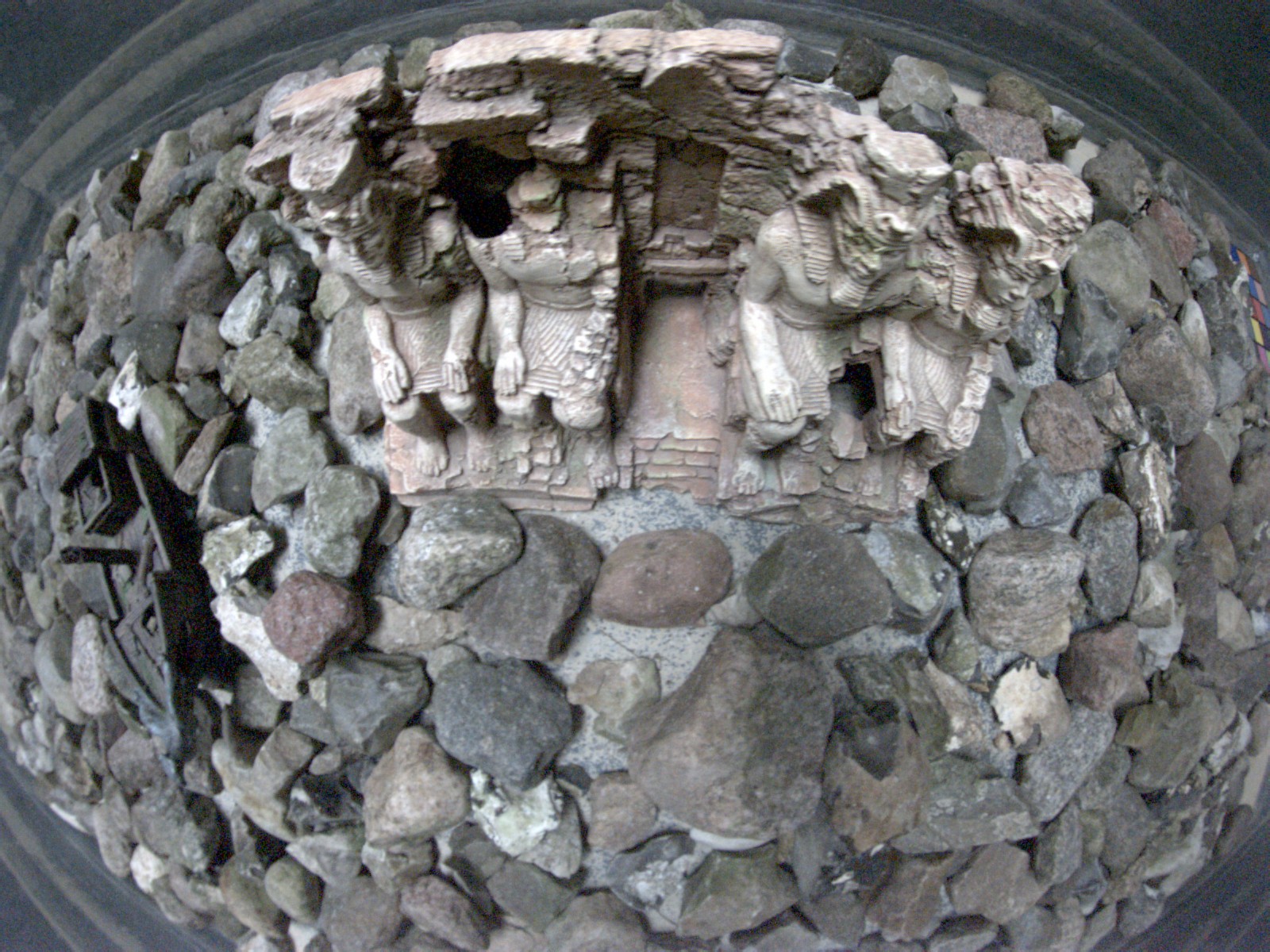}
       }
       \\~\\~\\
       \subfloat[Homogeneous sun illumination] {
           \includegraphics[width=0.33\textwidth]{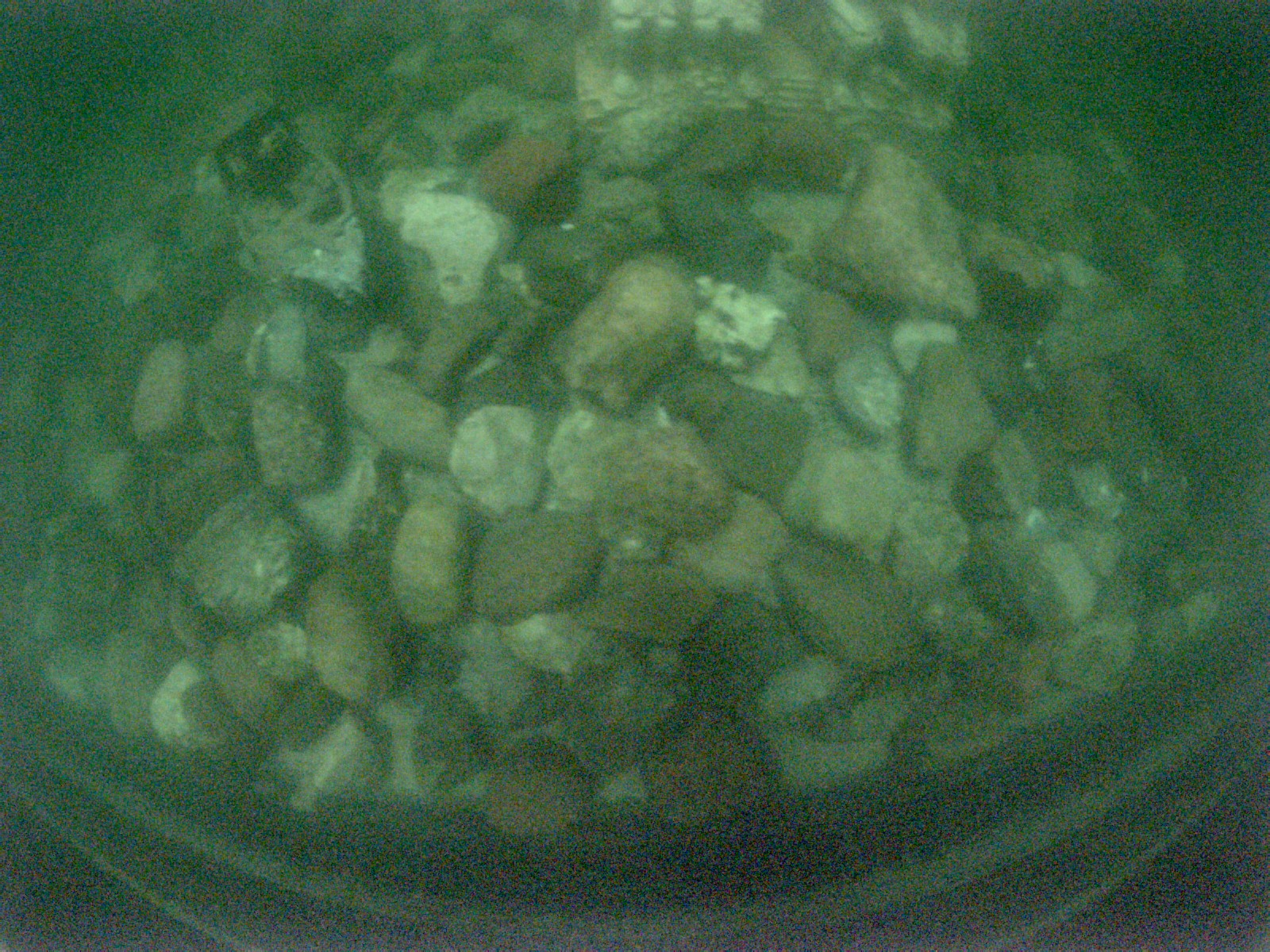}
           \includegraphics[width=0.33\textwidth]{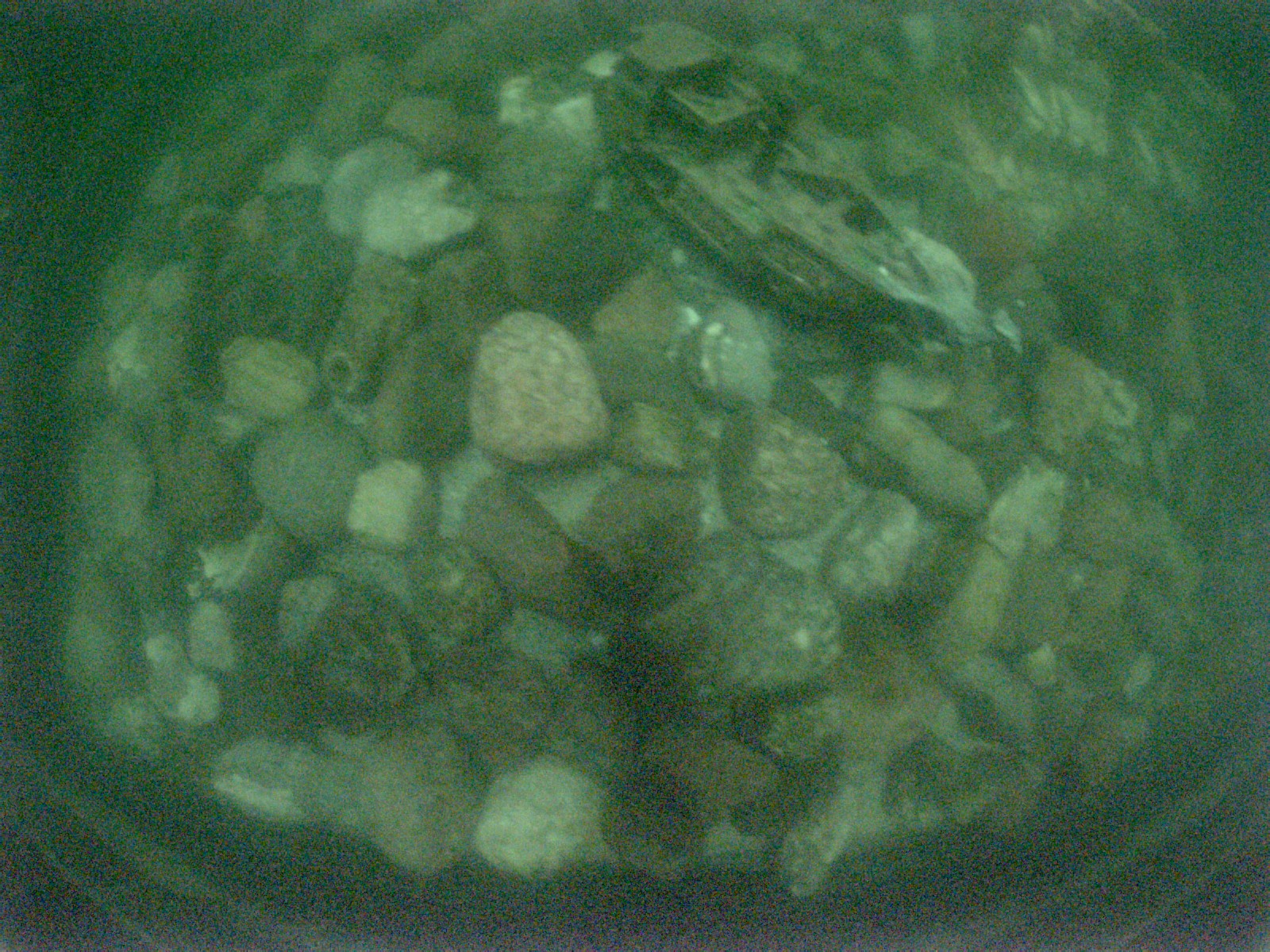}
           \includegraphics[width=0.33\textwidth]{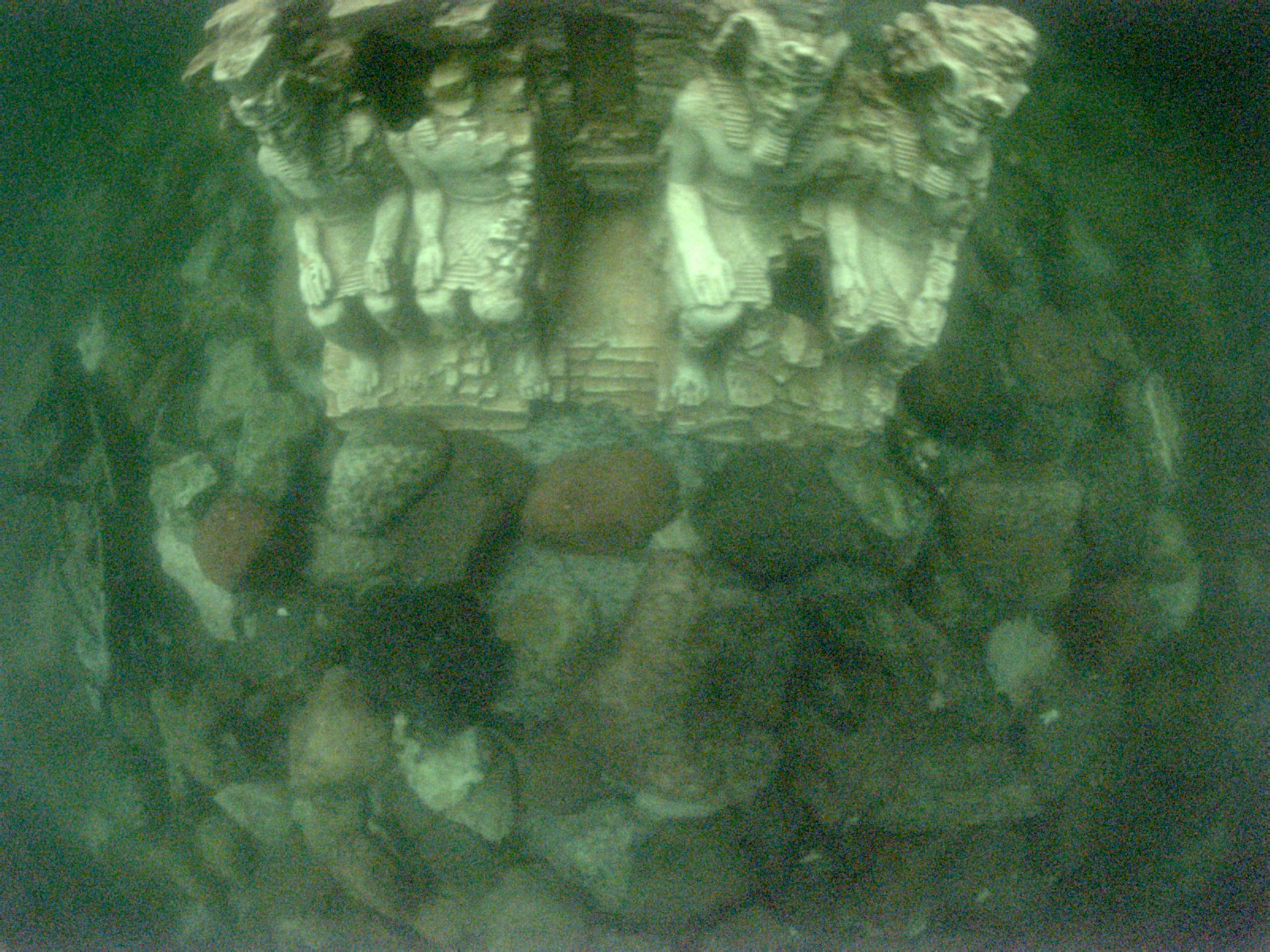}
       }
       \\~\\~\\
       \subfloat[Heterogeneous artificial illumination] {
           \includegraphics[width=0.33\textwidth]{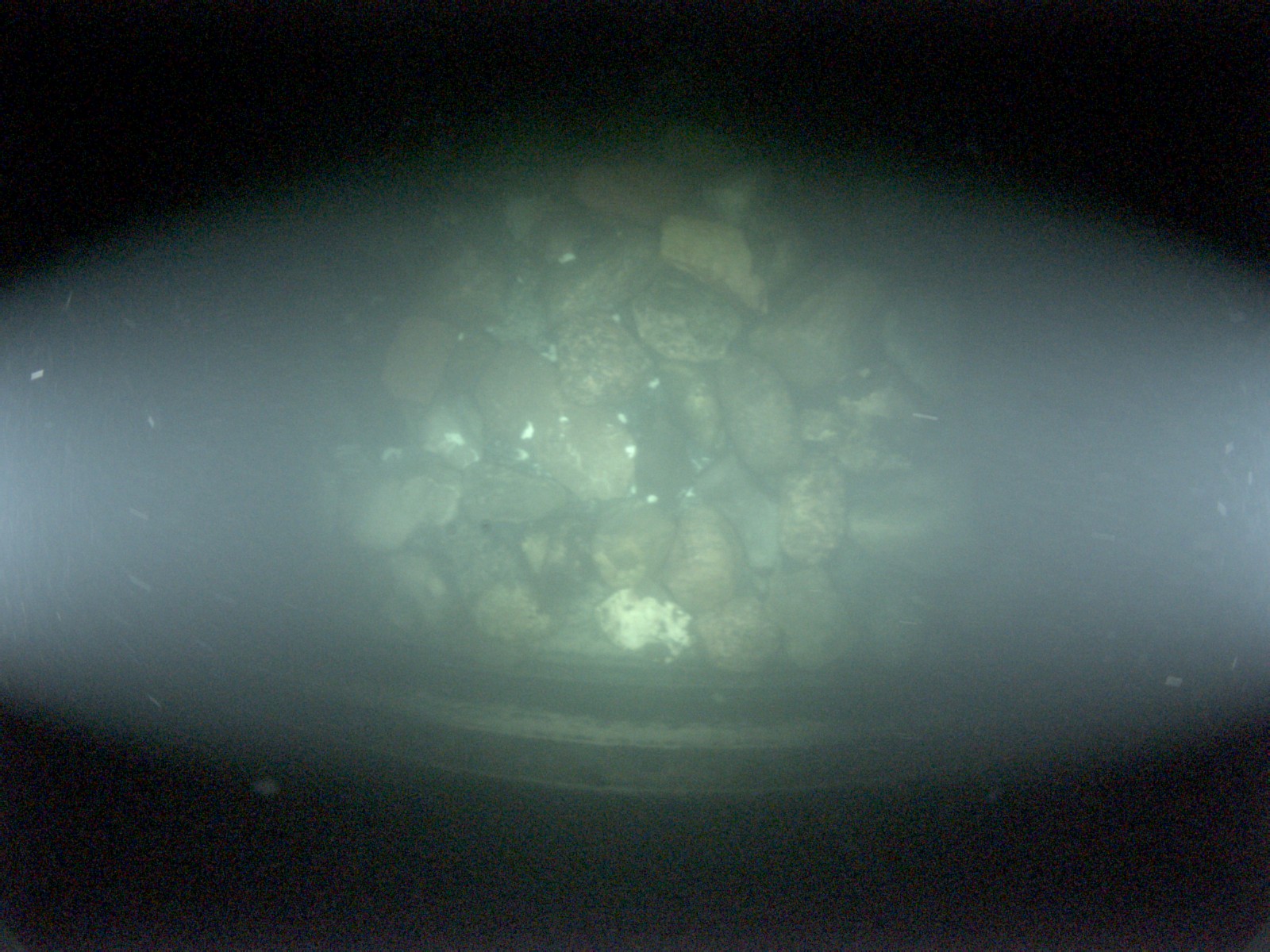}
           \includegraphics[width=0.33\textwidth]{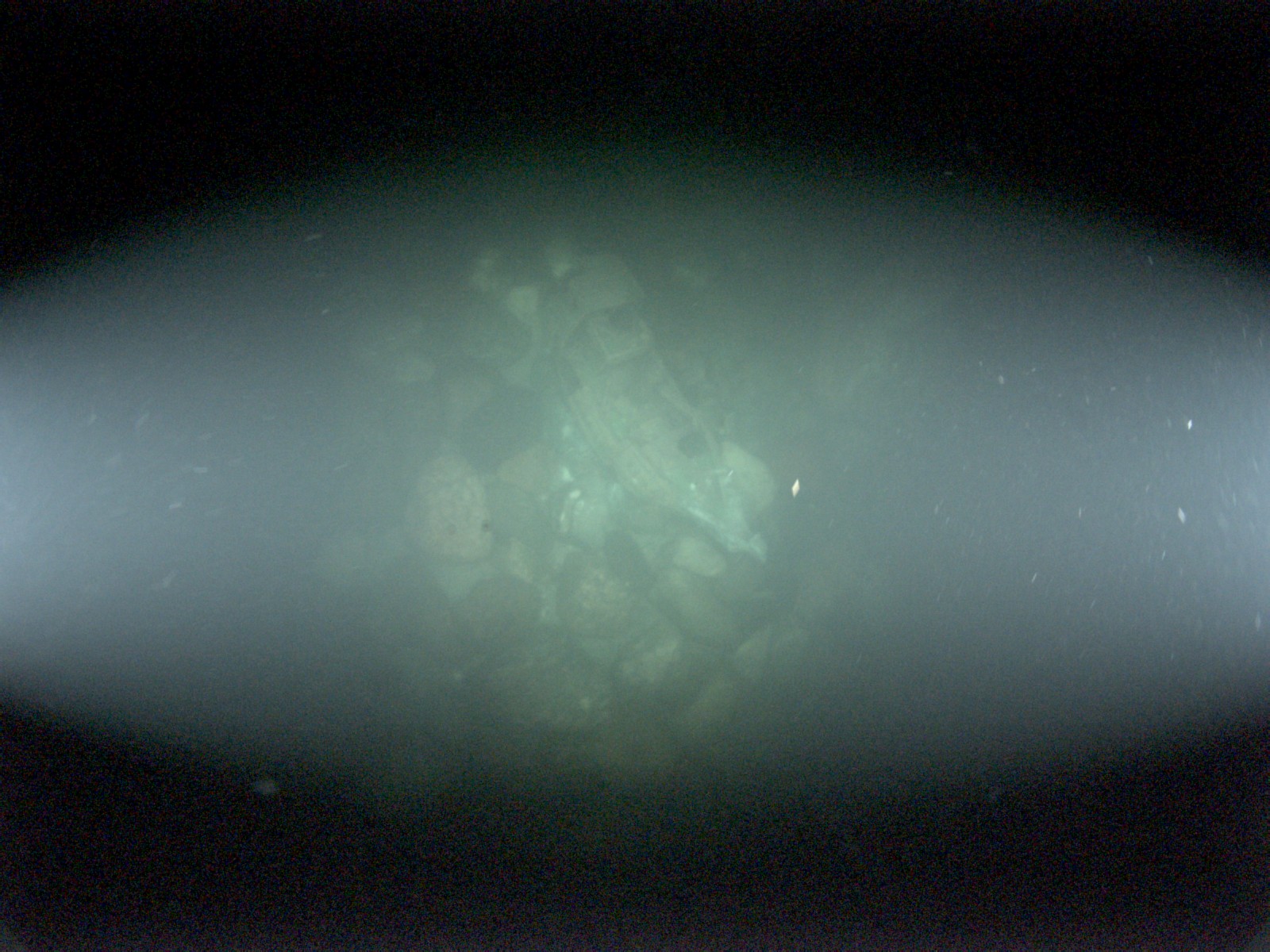}
           \includegraphics[width=0.33\textwidth]{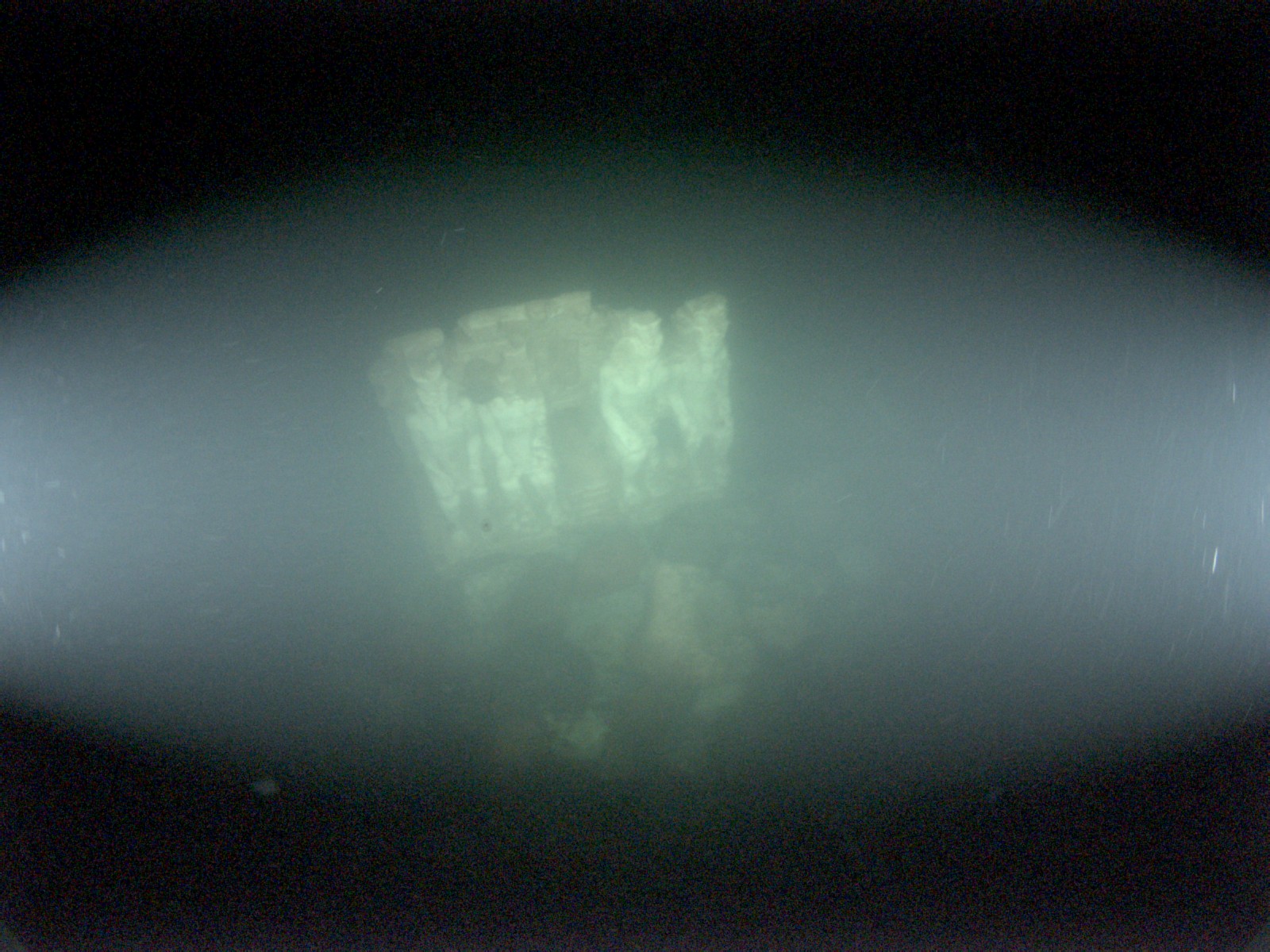}
       }
       \\~\\~\\
       \subfloat[Mixed (heterogeneous artificial and homogeneous sun) illumination] {
           \includegraphics[width=0.33\textwidth]{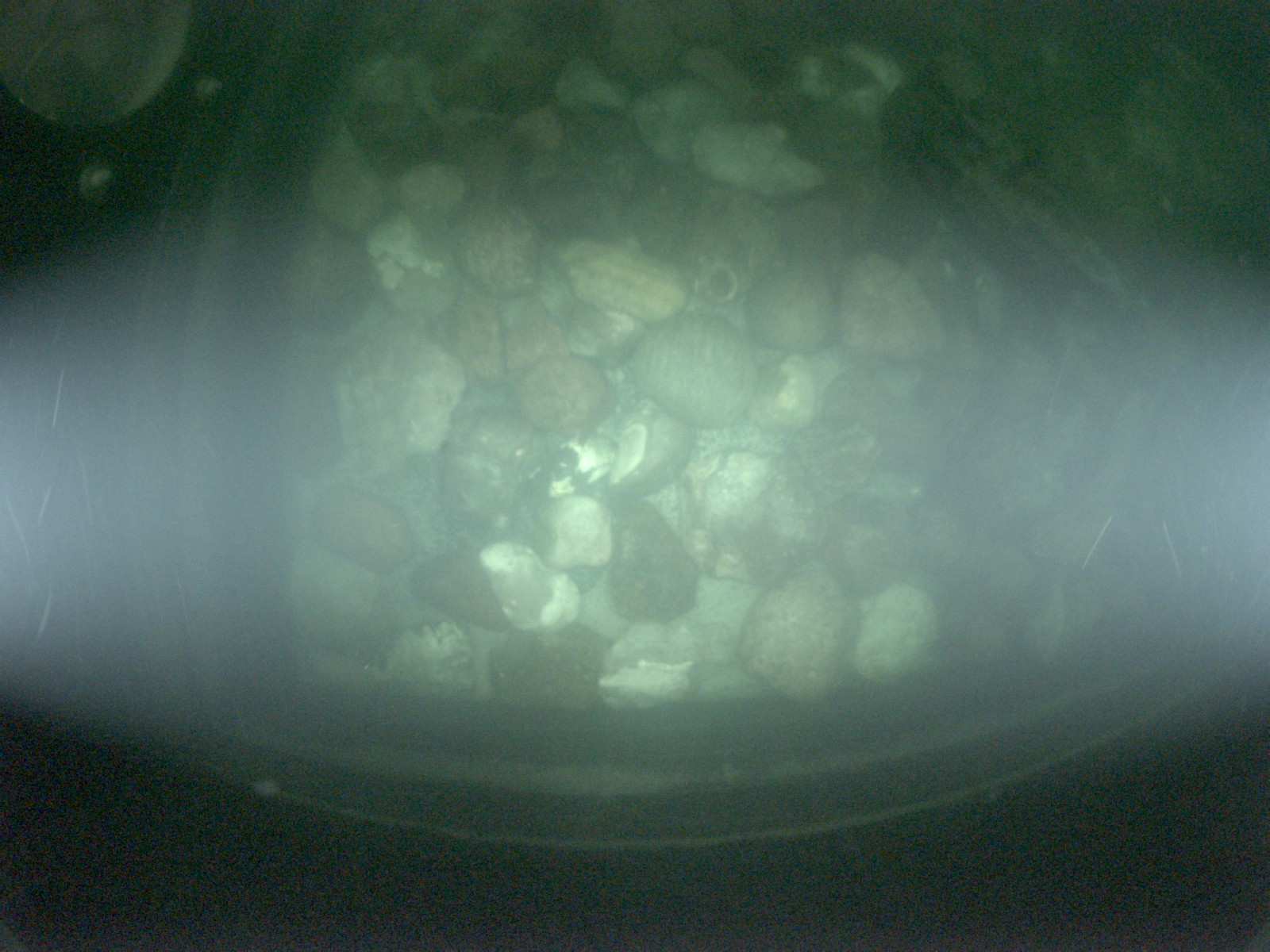}
           \includegraphics[width=0.33\textwidth]{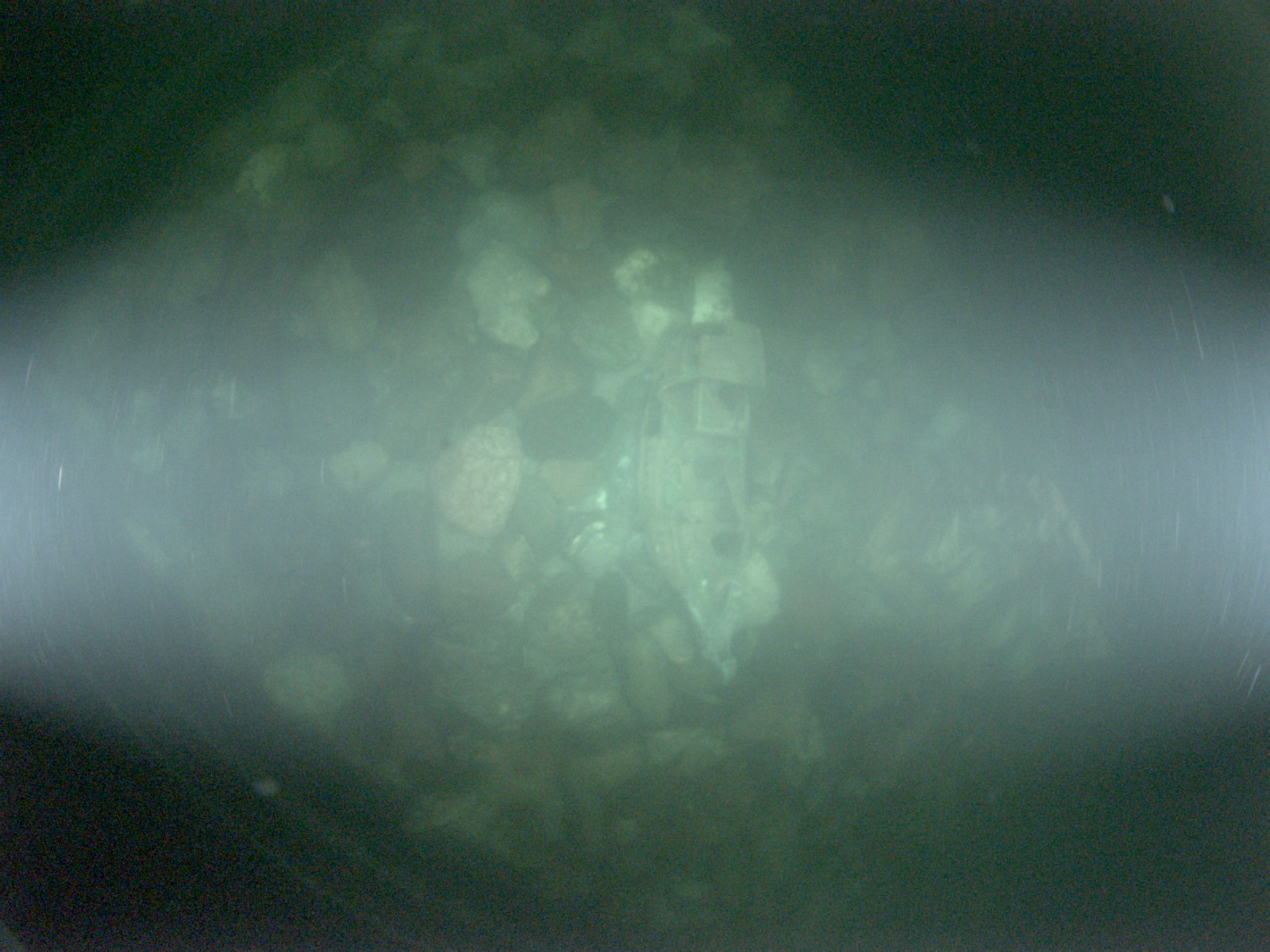}
           \includegraphics[width=0.33\textwidth]{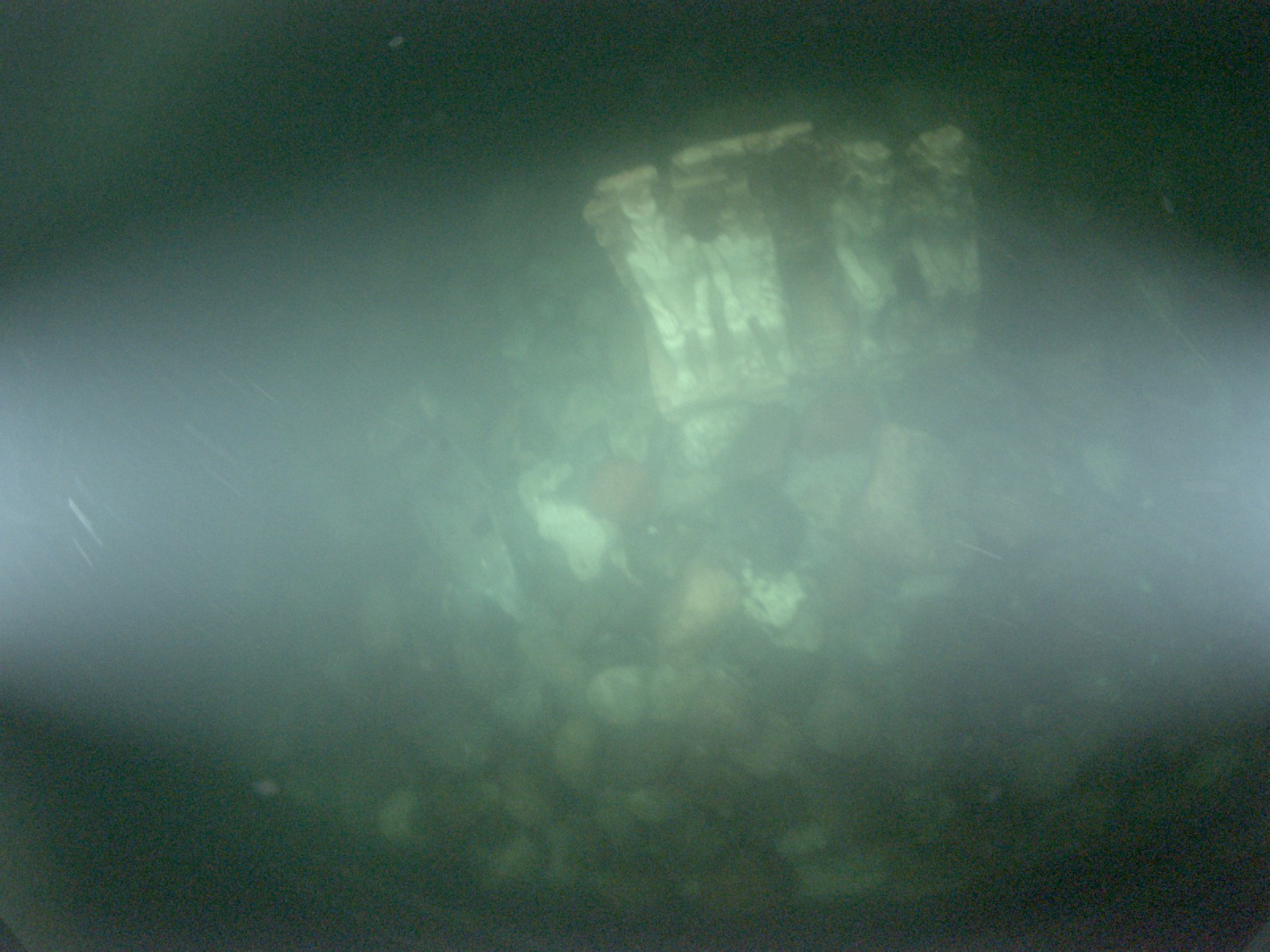}
       }
    \end{center}
    \caption{Captured underwater dataset. The RAW Images are still geometrically distorted but presented in sRGB space for better visibility.} \label{fig:uw_dataset}
\end{figure*}

\section{Conclusion}
\label{sec:conclusion}

In this paper we presented an external tracking system for underwater camera poses.
With the HTC Vive, we used an inexpensive consumer VR tracking system.
Given the tracking above the water surface and the result of the Hand-Eye calibration, a fused underwater camera pose can be estimated in real time.
Each system component is individually analyzed and validated.
Using the overall system, an average deviation of the camera pose from tracking to an optically determined camera pose of approximately 3 mm and 0.3° is achieved.
With the external reference system, ground truth poses are available for the underwater camera which enables comprehensive underwater experiments.

Future work will explore the possibility of an open-source tracking driver.
In addition, two IMUs will be mounted next to the camera.
One will be fused in the camera pose estimation, while the other will provide independent measurements which enables the evaluation on visual-inertial optical underwater methods.

\subsection{Future Application: Record Underwater Validation Datasets}

We recorded underwater datasets to proof the applicability of the system to the designated task (see \autoref{fig:underwaterExperiment}). 
To this end, we installed three 50W Wasler daylight (5400k) lamps with Walimex diffusors above the tank to mimic sunlight with heavy atmospheric scattering. 
In addition, we added two Ulanzi L2 lite (5500k) as co-moving light sources (see \autoref{fig:teaser}).
We added dye to the water to induce a seawater-like attenuation effect and in addition added Maaloxan as a scattering agent.

We now have a sensor-in-the-loop-system, which operates in a real scattering medium. 
Of course, a gap to the real world always remains, e.g, induced by missing swell, algae bloom, marine snow and the like. 
However, we trade this in for a tracking error that low that it cannot be achieved in the wild.

In \autoref{fig:uw_dataset} we show (a) in-air imagery and then the same scene under water:  with (b) homogeneous (sun) illumination, (c) heterogeneous illumination from co-moving lights and (d) a mix of sun and artificial illumination. 
In each scenario, we roughly captured 2000 images using traditional lawn mower patterns and a free 3D scanning scheme to capture the impact of depth variations on monocular underwater computer vision algorithms.
Each of the images then comes with a synchronized ground truth 3D pose, which e.g., was used in \cite{grimaldi2023investigation} to empirically investigate the difficulties of underwater visual monocular SLAM.
Finally, we fixed all parameters of the camera, except for exposure time and recorded RAW imagery. 
The linear nature of this data enables the development and testing of physically based color correction algorithms like e.g., \citep{https://doi.org/10.1002/rob.21638,akkaynak2019sea,nakath2021situ}.

Finally, our marker only features one ArUco marker per dimension.
In the future, this should be extended to more markers per dimension to yield an overdetermined system of equations for pose estimation.

\section*{Acknowledgement}
This publication has been funded by the German Research
Foundation (Deutsche Forschungsgemeinschaft, DFG) Projektnummer 396311425, through the Emmy Noether Programme.
The Authors would like to thank Marco Rohleder and Malte Eggersglü\ss ~from GEOMAR's Technology and Logistics Centre for providing support for designing and constructing the underwater camera housing.

\bibliographystyle{spbasic}      %
\bibliography{vive_tank_UW_tracking_minor_rev}   %

\end{document}